\documentclass[twoside]{article}
\usepackage{float}
\usepackage{algorithm}
\usepackage{algpseudocode}
\usepackage{makecell}
\usepackage{algpseudocode}
\usepackage[colorlinks=true, allcolors=blue]{hyperref}
\usepackage{enumitem}
\usepackage{graphicx}
\usepackage{subcaption}
\usepackage{amsmath, amssymb, amsthm}
\usepackage{multirow}
\usepackage[utf8]{inputenc} 
\usepackage[T1]{fontenc}    
\usepackage{hyperref}       
\usepackage{url}            
\usepackage{booktabs}       
\usepackage{amsfonts}       
\usepackage{nicefrac}       
\usepackage{microtype}      
\usepackage{xcolor}         
\usepackage{float}
\theoremstyle{plain}
\newtheorem{theorem}{Theorem}[section]

\newtheorem{lemma}[theorem]{Lemma}

\theoremstyle{definition}

\theoremstyle{remark}

\newcommand{\reals}{\mathbb{R}}

\usepackage[accepted]{aistats2026}
\usepackage[round]{natbib}

\begin{document}

%

%

\twocolumn[

\aistatstitle{{i-IF-Learn}: Iterative Feature Selection and Unsupervised Learning for High-Dimensional Complex Data}
\vspace{-0.5cm}
\aistatsauthor{ Chen Ma \And Wanjie Wang \And  Shuhao Fan }

\aistatsaddress{ \small SUSTech \And  NUS \And NUS } ]

\begin{abstract}
\vspace{-0.2cm}
Unsupervised learning of high-dimensional data is challenging due to irrelevant or noisy features obscuring underlying structures. It's common that only a few features, called the influential features, meaningfully define the clusters. Recovering these influential features is helpful in data interpretation and clustering. We propose {\it i-IF-Learn}, an iterative unsupervised framework that jointly performs feature selection and clustering. Our core innovation is an adaptive feature selection statistic that effectively combines pseudo-label supervision with unsupervised signals, dynamically adjusting based on intermediate label reliability to mitigate error propagation common in iterative frameworks. Leveraging low-dimensional embeddings (PCA or Laplacian eigenmaps) followed by $k$-means, i-IF-Learn simultaneously outputs influential feature subset and clustering labels. Numerical experiments on gene microarray and single-cell RNA-seq datasets show that i-IF-Learn significantly surpasses classical and deep clustering baselines. Furthermore, using our selected influential features as preprocessing substantially enhances downstream deep models such as DeepCluster, UMAP, and VAE, highlighting the importance and effectiveness of targeted feature selection. The implementation of i-IF-Learn is publicly available at
\url{https://github.com/mc25800852/i_if_learn}.
\end{abstract}
\vspace{-.35cm}
\section{INTRODUCTION}
\vspace{-.15cm}
Unsupervised learning, or clustering, is a fundamental learning task in various domains, including computer vision, natural language processing, and biomedical data analysis. 
Suppose $X_i \in \reals^p$ are observed, $1 \leq i \leq n$, each with a dimension of $p$. 
Clustering is to recover a label vector $\ell \in \{1, \cdots, K\}^n$, which reveals the hidden group structure of these $n$ data points.  Unlike supervised learning, there is no prior knowledge of $\ell$ and direct optimization methods are not available. 

Nowadays, researchers are facing challenges of complex data. For example, the feature dimension is much larger than the sample size, i.e., $p \gg n$, but 
many features may be irrelevant or even misleading for clustering. Simply applying a clustering algorithm on all dimensions often results in poor performance due to the curse of dimensionality \citep{curse, noise}. 
The features related to the intrinsic structure, which we call the influential features, are comparatively sparse. A feature selection step is critical. It identifies these influential features, which offers insights in the dataset and scientific problem. Furthermore, clustering methods can be applied on these influential features.

Even with a correct set of influential features, recovering the label vector $\ell$ remains difficult due to complex dependencies and low signal-to-noise ratios. Low-dimensional embedding methods  \citep{laplacian, UMAP, VAE} can extract the manifold structure and suppress noise. Incorporating these embeddings into the clustering step should enhance performance.

An iterative framework appears particularly promising, where estimated labels from previous iterations guide subsequent rounds of feature selection and clustering. However, iterative approaches face inherent challenges: early clustering errors may propagate, potentially reinforcing misleading patterns. Balancing the exploitation of early insights with safeguards against error propagation is thus essential.

\vspace{-.15cm}
\subsection{Related Work}
\label{sec:relatedwork}
\vspace{-.1cm}

Clustering has been widely studied, with classical algorithms such as $k$-means \citep{kmeans}, DBSCAN \citep{DBSCAN}, and spectral clustering \citep{SpecGEM}. While effective in low dimensions, these methods are sensitive to irrelevant features in high-dimensional data. To mitigate this, unsupervised feature selection approaches have been developed, see surveys \citep{li2017feature, zhao2019exploring}. For example, sparse $k$-means \citep{sparsekmeans, skmeans2} incorporates feature weighting, whereas Influential Features PCA (IFPCA) \citep{jin2017phase, ifpca} and IF Variational Auto-Encoder (IFVAE) \citep{chen2023subject} integrate feature screening with clustering.

Methods without explicit feature selection have also been proposed; see \cite{SC3, Seurat}. Examples include manifold fitting \citep{scAMF}, deep learning-based clustering methods \citep{DESC, svirsky2024interpretable}, and latent representation approaches \citep{VaDE}. These approaches focus on data reconstruction or latent representations. However, such methods cannot provide a direct understanding of the original features. Recent works address interpretability through differentiable feature selection \citep{FSLap, lee2022self, NeurCAM, SCTPC}, under the assumptions of dense features, semi-supervised settings, scenarios where $n \gg p$ or specific tpye of data. Furthermore, several advanced clustering frameworks \citep{Coper,itcluster, SCTPC} demand supplementary information.


Recently, there has been significant interest in iterative clustering frameworks. Deep clustering methods such as DEC \citep{DEC} and DeepCluster \citep{caron2019deepclusteringunsupervisedlearning} iteratively refine neural embeddings and cluster assignments, achieving strong empirical results in large-scale image and text datasets. However, these neural network-based methods often lack interpretability, limiting their utility in domains like genomics, where feature relevance insights are crucial. Methods like IDC \citep{svirsky2024interpretable} and CLEAR \citep{CLEAR} are also designed to improve interpretability, incorporated feature selection into clustering frameworks.

Outside the clustering context, feature selection in an unsupervised setting is also of great interest \citep{FSclustering, FSunsuperImage} to understand the data. 
Algorithms suggest that clustering labels could guide on feature selection \citep{FSkmeans, FSLap}. However, these methods perform feature selection independently from clustering, missing opportunities for iterative joint refinement.

\subsection{Our Contribution}

We propose the iterative Influential Features Learning ({\bf i-IF-Learn}) framework, explicitly designed for joint feature selection and clustering in high-dimensional, noisy datasets. Our approach integrates adaptive feature selection directly into the clustering pipeline, iteratively improving both clustering accuracy and feature interpretability. Our main contributions include:
\begin{itemize}
\item We introduce a novel composite statistic that adaptively balances supervised (pseudo-label-based) and unsupervised statistics for feature selection. 
Based on the reliability of the pseudo-labels, our statistic dynamically adjusting the reliance on them to mitigate error propagation in iterative clustering frameworks.
\item We develop two embedding-based clustering variants within our framework, i-IF-PCA and i-IF-Lap, employing PCA and Laplacian eigenmaps, respectively. Empirical results indicate superior performance of nonlinear embeddings (i-IF-Lap), highlighting the benefits of capturing complex manifold structures.
\item Our method simultaneously outputs cluster labels and an interpretable set of influential features. This feature subset substantially enhances downstream analyses, improving the clustering performance of state-of-the-art methods like DeepCluster, UMAP, and VAE when used as preprocessing.
\item We establish consistency results for both label recovery and feature selection under a weak signal model. Experiments on microarray and single-cell RNA-seq datasets show superior performance over classical and deep clustering methods. Furthermore, i-IF-Lap+deep clustering methods leads to significant improvements.
\end{itemize}

\begin{figure}[H]
    \includegraphics[width=\linewidth]{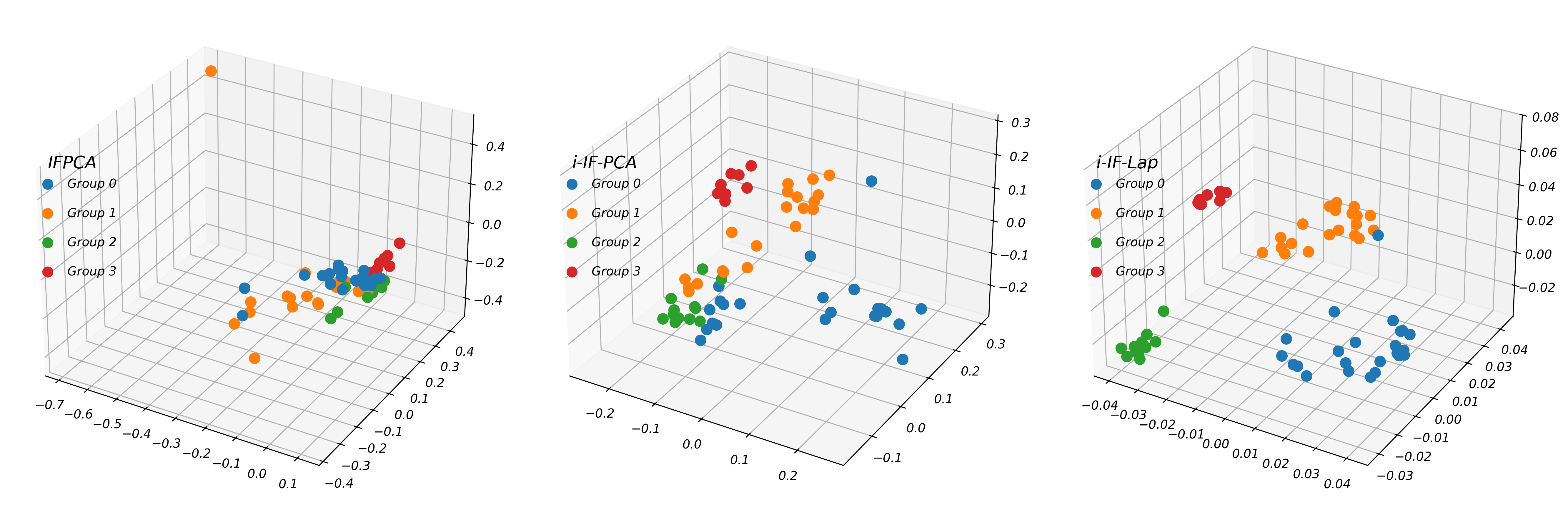}
    \caption{Eigenvector projection of data points in 3D space using three clustering pipelines. }
    \label{fig:3d_pca_cluster}
\end{figure}
To illustrate the effect of iteration and embedding choice, we analyze SRBCT dataset using three methods: (a) IFPCA, a non-iterative baseline\citep{ifpca}; (b) i-IF-PCA, our iterative variant using PCA; and (c) i-IF-Lap, using Laplacian eigenmaps for nonlinear embedding. Figure  \ref{fig:3d_pca_cluster} displays the resulting 3D embeddings by these methods, colored by ground-truth labels. With the iteration step, both i-IF-PCA and i-IF-Lap have a better embedding than IFPCA. Further, due to the nonlinear property of gene microarray data, i-IF-Lap using a Laplacian eigenmap embedding performs better than i-IF-PCA. 
The comparison shows that both iteration and nonlinear embeddings substantially improve cluster separation. More comprehensive numerical results can be found in Sections \ref{sec:data} and \ref{sec:simu}.

\vspace{-.2cm}
\subsection{Organization}
\vspace{-.2cm}
We first introduce our i-IF-Learn framework in Section \ref{sec:algo} with technical details. In Section \ref{sec:model}, we introduce the model and theoretical results. Numerical results can be found in Section \ref{sec:data} on real data sets and Section \ref{sec:simu} on synthetic data. Proofs, implementation details, and data description are left to the appendix. 

\vspace{-.2cm}
\section{ALGORITHM: ITERATIVE HIGH-DIMENSIONAL CLUSTERING}\label{sec:algo}
Consider the data matrix $X \in \reals^{n \times p}$, where each row is $X_i \in \reals^p$ with a high dimension $p$. 
For each data point, the label is denoted as $\ell_i \in [K]$ for a known constant $K$, where $K$ is the number of clusters. 
Denote $I$ to be the set of these influential features, where 
$ I = \{1\leq j \leq p: E[X_{ij}|\ell_i = k_1] \neq E[X_{ij}|\ell_i = k_2], \mbox{ for some $1\leq k_1 \neq k_2 \leq K$}\}$, $k_1$ and $k_2$ represent two arbitrary distinct cluster labels in the set $\{1,\dots ,K \}$. It means if there are two clusters with different expectations on this feature, then this feature if influential feature. We aim to recover both the latent cluster labels $\ell$ and the set of influential features $I$ that drive the clustering structure.

\begin{figure}[H]
    \centering
    \includegraphics[width=\linewidth]{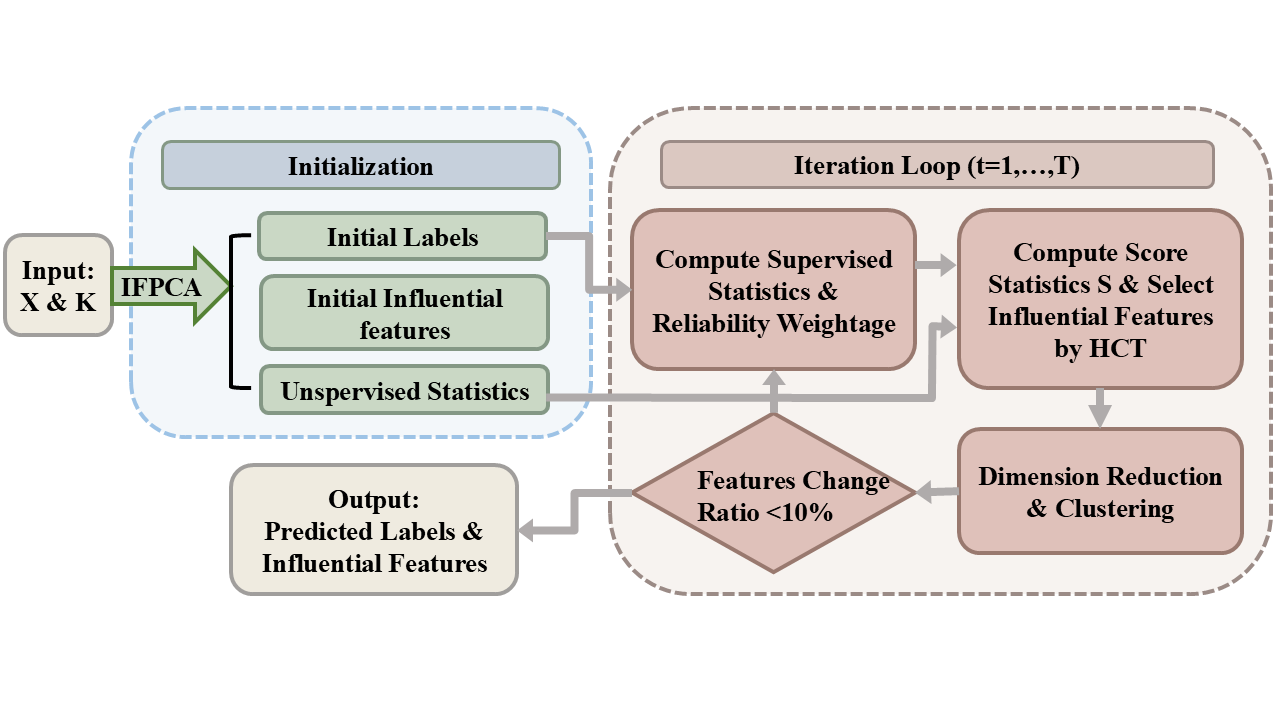}
    \caption{Overview of the i-IF-Learn framework. }
    \label{fig:workflow}
\end{figure}

We propose the i-IF-Learn algorithm, an iterative clustering framework with an initialization stage and an iterative refinement loop, as illustrated in Figure \ref{fig:workflow}. The initialization stage recovers relatively strong signals with a noisy clustering label, and the iterative loop further recovers the weak signals and refines the clustering assignments. It is outlined as Algorithm \ref{alg:i-if-learn}. In Sections~\ref{sec:ifpca}--\ref{sec:stop}, we explain the idea of each step. The complete algorithm with every implementation detail can be found in Algorithm \ref{alg:iiflearn} in Appendix.

\begin{algorithm}[h]
\caption{i-IF-Learn Algorithm}
\label{alg:i-if-learn}
\textbf{Require:} Data matrix $X \in \reals^{n \times p}$, number of clusters $K$, maximum iterations $T$ \\
\textbf{Ensure:} Clustering labels $\hat\ell$, selected features $\hat{I}$.

\textbf{Stage 1: Initialization} \\
\textbf{Require:} Data $X \in \reals^{n \times p}$, number of clusters $K$\\ 
\textbf{Ensure:} Initial pseudo-labels $\ell^{(0)}$ and feature set $I^{(0)}$

\textbf{Stage 2: Iterative Loop}
\begin{algorithmic}[1]
    \For{$t = 1, 2, \dots$}

        \State \textbf{Step 1:} For each feature $j$, compute score $S_j^{(t)}$  
        \Statex \hspace{1.2em} based on $\ell^{(t-1)}$, set Higher Criticism threshold  
        \Statex \hspace{1.2em} $\tau^{(t)}$, and update $I^{(t)} = \{1 \le j \le p : S_j \ge \tau^{(t)}\}$.

        \State \textbf{Step 2:} Construct the post-selection data  \Statex \hspace{1.2em} matrix $X^{(t)} = X^{(I^{(t)})}$, perform low-dimensional 
        \Statex \hspace{1.2em} embedding to obtain $U^{(t)}$, and run $k$-means on 
        \Statex \hspace{1.2em} $U^{(t)}$ to get $\ell^{(t)}$.

        \State \textbf{Step 3:} Compute influential feature change 
        \Statex \hspace{1.2em} ratio $r = r(I^{(t-1)}, I^{(t)})$.
        \If{$r \leq 10\%$ or $t = T$}
            \State \textbf{break}
        \EndIf
    \EndFor
\State $\hat{\ell} = \ell^{(t)}$, $\hat{I} = I^{(t)}$. 
\end{algorithmic}
\end{algorithm}

In IF step, due to the large $p$, we employ a ranking and thresholding procedure for computation efficiency. 
For each feature $j$, we propose a composite score 
\begin{equation}
    S_j^{(t)} = \omega^{(t)} T_j^{sup}(\ell^{(t-1)}) + (1-\omega^{(t)})\, T_j^{unsup},
\end{equation}
where $T_j^{sup}(\ell^{(t-1)})$ is a pseudo-label-supervised test statistic and $T_j^{unsup}$ is an unsupervised statistic. 

The weight for the supervised statistic is defined as
\begin{equation}\label{equ:weight}
    \omega^{(t)} = \frac{w^{(t)}}{\sqrt{(w^{(t)})^2 + (1 - w^{(t)})^2}},
\end{equation}
where $w^{(t)}$ reflects the reliability of the label estimates $\ell^{(t-1)}$, 
and $(1-\omega)$ correspondingly represents the weight assigned to the unsupervised statistic. 

We then compute $p$-values based on $S_j^{(t)}$ and apply Higher Criticism thresholding (HCT) to obtain the updated influential feature set $I^{(t)}$. 
This procedure is entirely data-driven and requires no tuning parameters; details are in Sections~\ref{sec:stat}--\ref{sec:HCT}.

In the Learn step, we perform low-dimensional embedding of $X^{(I^{(t)})}$ on the selected features $I^{(t)}$. 
The algorithm has variants depending on the selection of embedding methods. We call it {\bf i-IF-PCA} when PCA is employed for embedding and {\bf i-IF-Lap} when Laplacian eigemap is used for embedding. 
To capture the low-dimensional manifold structure, i-IF-Lap usually performs better. 
$K$-means clustering is then applied to the embedded data to generate updated labels $\ell^{(t)}$. 

The algorithm continues until the feature set $I^{(t)}$ stabilizes or a maximum number of iterations is reached. The final outputs are the estimated labels $\hat{\ell} = \ell^{(T)}$ and influential feature set $\hat{I}=I^{(T)}$. 
The initialization step has a computation cost of $O(n + ns)$ where $s$ is the number of selected features. 
For every iteration, the complexity for $S_j$ is $O(n)$ and for clustering is $O(n^2s)$. Therefore, the overall computation complexity is $O(Tn^2s)$, where $T$ is the number of iterations, $n$ is the sample size and $s$ is the number of selected features.  

By the iterative framework, i-IF-Learn recovers features with both the relatively strong signals and weak signals. To illustrate the effectiveness of $\hat{I}$, we apply multiple clustering methods to data restricted on $\hat{I}$ to get the cluster labels, instead of the estimated labels by i-IF-Lap itself. 
In numerical analysis, we consider UMAP, DeepCluster, and VAE as the downstream clustering methods, and all of them have a significant improvement in clustering. It further proves that our algorithm provides interpretable results that deepens our understanding of data.

\subsection{Initialization}\label{sec:ifpca}

The i-IF-Learn algorithm begins with an initialization step to estimate the initial cluster labels $\ell^{(0)}$ and the influential feature set $I^{(0)}$. 
While our theoretical guarantees allow for any reasonable initialization, we adopt the IFPCA method \citep{ifpca} for two main reasons. 
First, IFPCA has demonstrated strong theoretical and empirical performance in high-dimensional clustering. It reliably identifies features with relatively strong signals and produces stable clustering labels. Second, 
our iterative procedure leverages both $T^{sup}$ and $T^{unsup}$. The unsupervised component $T^{unsup}$ coincides with the IF step in IFPCA, reducing the computation cost. 
The IFPCA procedure is summarized in Algorithm \ref{alg:ifpcashort}, with implementation details deferred to Algorithm \ref{alg:ifpca} in Appendix.

\begin{algorithm}
\caption{IFPCA Initialization Procedure}
\label{alg:ifpcashort}
\begin{algorithmic}[1]
\Require Data $X \in \reals^{n \times p}$, number of clusters $K$
\Ensure Initial labels $\ell^{(0)}$, influential feature set $I^{(0)}$

\State {\bf Step 1}: Compute unsupervised test scores
\Statex $\psi_{n,j} \gets$ Kolmogorove-Smirnov score between the empirical CDF of $x_j$ and normal CDF. 

\Statex Normalize scores: $\psi^*_{n,j} \gets \frac{\psi_{n,j} - \text{mean}(\psi_{n,\cdot})}{\text{std}(\psi_{n,\cdot})}$

\State {\bf Step 2}: Feature selection by HCT
\Statex $\pi_j \gets 1 - F_0(\psi^*_{n,j})$, where $F_0$ is the null distribution.
\Statex $HC_{j} \gets$ a function based on $\pi_j$
\State $\hat{j} \gets \arg\max_j HC_{p,j}$,  $t_p^{\text{HC}} \gets \psi^*_{n, \hat{j}}$
\Statex Selected features: $I^{(0)} \gets \{1\leq j \leq p \mid \psi^*_{n,j} > t_p^{\text{HC}} \}$

\State {\bf Step 3}: PCA embedding and $k$-means clustering
\Statex Apply PCA to post-selection data, retain top $K-1$ components. Denote it as $U$.
\Statex Labels: $\ell^{(0)} \gets$ $k$-means({$U$, $K$})

\end{algorithmic}
\end{algorithm}

\subsection{Novel Iterative Screening Statistic}\label{sec:stat}

A key ingredient in the IF step is the score statistic $S_j^{(t)}$, which determines how relevant each feature is to the underlying structure. 
Unlike traditional screening methods that rely solely on either supervised or unsupervised tests, 
we introduce a new composite statistic that adaptively combines both sources of information as follows:

\vspace{-0.1mm}
\begin{equation}\label{eqn:s}
\hspace{-0.1cm}
   S_j^{(t)} = \omega^{(t)} \Phi^{-1}(1 - P_{F, j}^{(t)}) + (1-\omega^{(t)})\Phi^{-1}(1 - P_{KS, j}).
\end{equation}

Here, $T^{sup}_j(\ell^{(t-1)}) = \Phi(1 - P_{F, j}^{(t)})$ denotes the supervised test statistic and $T^{unsup}_j = \Phi^{-1}(1 - P_{KS, j})$ denotes the unsupervised test statistic. 
In detail, $P_{F, j}^{(t)}$ is the $p$-value of marginal $F$-statistic, using $x_j$ and the current label estimates $\ell^{(t-1)}$. 
$P_{KS, j}$ is the $p$-value of the Kolmogorov-Smirnov (KS) statistic between the empirical distribution of $x_j$ for feature $j$ and a specified null distribution \citep{KS}. While our framework is general and can accommodate any appropriate null distribution based on the specific domain, in this work we adopt the standard normal distribution as the null. This choice aligns with the rare and weak signal setting commonly assumed for high-dimensional genetics data \citep{ifpca}.
$P_{KS, j}$ remains static among iterations while $P_{F, j}^{(t)}$ depends on the pseudo-label in every iteration. Both are corrected to get rid of the gap between empirical null and theoretical null \citep{EmpNull, ifpca}. 
The KS statistic $P_{KS, j}$ captures distributional deviation, while the F-statistic $P_{F, j}^{(t)}$ measures separation across pseudo-label clusters. Their associated $p$-values provide a statistically grounded measure of feature importance.
Finally, we transform them into normal quantiles to calculate $S_j^{(t)}$, instead of using $p$-values \citep{wang2022sparse} or the original statistics. 
Hence, the selected features are interpretable. 

The \textbf{reliability weightage} $w^{(t)} \in [0,1]$ is to evaluate our trust in the current estimated label $\hat{\ell}^{(t-1)}$. When the pseudo-label $\ell^{(t)}$ is more reliable, we tend to have a larger $w^{(t)}$ on the supervised statistic, with a higher power in feature selection. 
However, without the ground-truth of labels, the trust in $\ell^{(t)}$ is difficult to evaluate. 

The idea is, if the selected features in previous step $l^{(t-1)}$ is further away from noises, then the predicted label $\ell^{(t)}$ based on $l^{(t-1)}$ should be more reliable. Consider the set of $p$-values from $F$-statistics $\{P_{F,j}; j \in I^{(t-1)}\}$. 
We conduct a hypothesis testing on whether $P_{F,j}$ contains information or not, using this set. Let $p_1^{(t)} \in (0, 1)$ denote the $p$-value of this test, where the details can be found in Appendix \ref{sec:weight}. A smaller $p_1^{(t)}$ indicates a larger possibility that $I^{(t-1)}$ is informative. 
The definition of $w^{(t)}$ as follows: 

\begin{equation}\label{eqn:w}
    w^{(t)} = 1- {p_1^{(t)}}/{ (p_1^{(t)} + c)}. 
\end{equation}

The constant $c$ gives the default importance of $P_{F,j}$, at $c/(1+c)$. 
Even when $p_1^{(t)} \to 1$, i.e., the beginning stage, we still hope $P_{F,j}$ to take part in. When $p_1^{(t)} \to 0$, the weight gradually approach $1$, no matter what $c$ is. A reasonable range for $c$ is $0.35 \leq c \leq 0.6$. In numerical analysis, we consistently use $0.6$, with an experiment on the effects of $c$ in Section \ref{sec:simu}.

\subsection{Feature Selection by HCT}\label{sec:HCT}

Our novel score statistic $S_j^{(t)}$ ranks the importance of features, where a larger score $S_j^{(t)}$ indicates a larger potential for feature $j$ to be influential. Hence, the selection follows that, for a threshold $\tau^{(t)}$, 
\begin{equation}\label{eqn:selection}
    I^{(t)} = \{1 \leq j \leq p \mid S_j^{(t)} \geq \tau^{(t)}\}. 
\end{equation}
The key is to decide $\tau^{(t)}$ in every iteration. We apply HCT, a data-driven threshold that optimizes the selection.  First derive the $p$-values $\pi_j^{(t)} = 1 - \Phi(S_j^{(t)})$, then order the $p$-values in an increasing order, $\pi_{(1)}^{(t)} \leq \pi_{(2)}^{(t)} \leq \cdots \leq \pi_{(p)}^{(t)}$. The HCT is defined as 

\begin{eqnarray}\label{eqn:HCT}
    \tau^{(t)} = S_{\hat{j}}^{(t)}, 
\end{eqnarray}
 $$\mbox{where } \hat{j} = \arg\max_{\log p \leq j \leq p/2} {(j/p - \pi_{(j)}^{(t)})}/{\sqrt{\pi_{(j)}^{(t)}(1 - \pi_{(j)}^{(t)})}}$$

Using this HCT in \eqref{eqn:selection}, the $\hat{j}$ features with  largest scores, i.e. smallest $p$-values, are selected. The selected features are considered as the influential features.

The corresponding post-selection data matrix follows that $X^{(t)} = X[:, I^{(t)}]$.

\subsection{Embedding and Clustering}\label{sec:Learn}

After selecting influential features, we normalize the corresponding submatrix $X^{(t)}$ to obtain $W(t)$, where each column has mean zero and unit variance. Despite feature selection, the high-dimensional data still contain redundant noise. Thus, we apply low-dimensional embedding techniques to extract underlying structure and enhance the signal-to-noise ratio.

 We consider two embedding methods: Principal Component Analysis (PCA) and Laplacian Eigenmaps. PCA \citep{PCA, wang2022sparse, uniformpca} projects the data onto directions of maximum variance and is widely used due to its simplicity and interpretability. However, it may fail to capture complex nonlinear structures inherent in many modern datasets. 
 Laplacian Eigenmaps \citep{laplacian}, in contrast, construct a data dissimilarity graph and compute embeddings in to spectral space that preserve local geometry. This approach is particularly effective when data lie on a nonlinear manifold. Other embedding methods, such as UMAP \citep{UMAP} and Autoencoder \citep{10.5555/3045796.3045801}, are also evaluated; their comparative results on real data sets can be found in the Appendix \ref{sec:app-addtional}.

For both methods on $W^{(t)}$, we consider the embeddings into $K+2$-dimensional space. Under the low-rank structure, separating $K$ clusters only requires an embedding dimension larger than $K-1$ \citep{DudaHartStork2001Chap5}. In this work, we conservatively choose $K+2$ dimensions to ensure that the embedding preserves sufficient structural information. Therefore, we construct a spectral matrix $U^{(t)}$ from either $W^{(t)}$ or the data similarity matrix: 

\begin{equation}
    U^{(t)}=[u_1^{(t)},u_2^{(t)},\dots,u_{K+2}^{(t)}] \in \reals^{n\times(K+2)}.
\end{equation}


Then we perform $k$-means on $U^{(t)}$, treating each row as a data point, to obtain pseudo-labels $\ell^{(t)} = k\text{-means}(U^{(t)},K)$.

In real data analysis, we find that both embedding methods outperform clustering on raw features, with Laplacian Eigenmaps (i-IF-Lap) consistently yielding better results, highlighting the utility of nonlinear embeddings in complex data.

\subsection{Stopping Criteria}\label{sec:stop}

To determine convergence, we monitor the stability of the selected influential feature sets across iterations. Let the relative change rate between iterations $t-1$ and $t$ be 
$r^{(t)} = {\left| I^{(t)}/I^{(t-1)} \right|}/{\left| I^{(t-1)} \right|}$. If the change rate $r^{(t)} \leq 10\%$ or the number of iterations exceeds a fixed limit (e.g., 10), we terminate the process.

\section{MODEL ASSUMPTIONS AND THEORETICAL GUARANTEE}\label{sec:model}

Consider the asymptotic clustering model where the signals are rare and weak. In detail, the data matrix $X_i \sim N(\mu_0 + \mu_{\ell_i}, \Sigma)$, where $\Sigma$ is a diagonal matrix with diagonals $\sigma^2_j$. Let $M = [m_1, m_2, \cdots, m_K] \in \reals^{p\times K}$, where $m_k = \Sigma^{-1/2}\mu_k$ be the normalized mean vectors. Let $M_j$ denote the $j$-th column of $M$. 
 The influential features set is that 
$    I = \{1 \leq j \leq p \mid \|M_j\| \neq 0\}$. 
Asymptotically, we assume that the signals are sparse, in the sense that when $p \to \infty$, 

\vspace{-0.8mm}
\begin{equation}\label{eqn:rw}
    \epsilon = {|I|}/{p} \to 0.
\end{equation}
The sparse setting causes a low signal-to-noise ratio, which is challenging in high-dimensional unsupervised learning. 

While most theoretical papers focus on a uniform signal strength to decide the detection boundary, it is not the case in practice. The signal strength in individual features has severe heterogeneity. 
By an iterative framework, based on the extremely sparse features with relatively strong signals, we have an initial clustering label, and then recover the features with weak signals in iterations. 

The following theorem explains the iterative feature selection effect. 
The technical conditions are not strict. In the simplified scenario that $K = 2$ with equal group size, as long as the accuracy rate is a constant larger than $1/2$, then the condition is satisfied. 
\begin{theorem}\label{thm:fs}
    Consider the estimated label $\hat{\ell}$, which can be the initial label $\hat{\ell}^{(0)}$ or the estimated label $\ell^{(t-1)}$ from last round. Denote $w_{ij} = E[\Sigma^{-1/2}X_{ij}]$ as the expectation for data point $i$ on feature $j$, and the overall mean $\bar{w}_j=\frac{1}{n}\sum_{i=1}^n w_{ij}$. Denote $n_k(\hat{\ell}) = \sum_{i=1}^n 1\{\hat{\ell}_i = k\}$, $k \in [K]$. 
    Suppose for the influential features in $I$, the community label satisfies that for a constant $c_0 > 0$, 
\vspace{-5mm}

\begin{equation}
    \min_{j\in I} U_j\geq  c_0(\log p)^2,
\end{equation}   
\vspace{-6mm}

$$\mbox{where } U_{j}:=\sum\nolimits_{k\in[K]}\frac{1}{n_k(\hat{\ell})} \sum\nolimits_{\hat{\ell}_i = k}(w_{ij} -\bar{w}_j)^2. $$
Then our weight selection satisfies $P(w^{(t)} \geq 1-p^{-2})\geq 1-p^{-2}$. Furthermore, with probability $1-p^{-4}$, the $\hat{I}$ from IF step in i-IF-Learn satisfies that 
$I\subset\hat{I}$ and $|\hat{I}/I|\leq C_0 (\log p)^2$. 
\end{theorem}

The correct recovery of $I$ leads to a correct label recovery, as long as $I$ contains sufficient information. 
\begin{theorem}\label{thm:clustering}
Suppose the assumptions of Theorem \ref{thm:fs} hold. Let $\hat{I}$ denote the selected feature set and $M_{\hat{I}}$ denote the mean feature matrix $M$ restricted on $\hat{I}$. 
Let $\tau_{\hat{I}}$ be the eigengap of matrix $M_{\hat{I}}$, then with a high probability, the $K$-means clustering error follows that 

\begin{equation}
    Err(\hat{\ell}, \ell) \leq C{({n^{1/2}} + {|\hat{I}|}^{1/2})^2}/{n\tau_{\hat{I}}^2}. 
\end{equation}

Further, when the signal strength satisfies $\min_{j \in I} \|M_j\| \geq \log^2 p/\sqrt{n}$, then the clustering label by i-IF-Learn has that 
$Err(\hat{\ell}, \ell) \to 0$.
\end{theorem}

Our theorems suggest that when $p$ is sufficiently large, one-step iteration could recover the correct labels and influential feature set from a random initialization. In practical data, since the constants are difficult to decide and $p$ might be inadequate, we run iterations multiple times to ensure robustness and enhance practical performance. 

\begin{table*}[h]
\centering
\caption{Accuracy comparison of clustering methods across 10 gene microarray datasets.}
\label{acc1}
\resizebox{\textwidth}{!}{
\begin{tabular}{lcccccccccccc}
\toprule
\textbf{Dataset} & KMeans & SpecGEM & UMAP & DEC & DeepCluster & IFPCA & IFVAE & i-IF-PCA & i-IF-Lap & i-IF-Lap+UMAP & i-IF-Lap+DeepCluster & i-IF-Lap+VAE\\
\midrule
Brain     & 0.667 & \textbf{0.857} & 0.676 (0.07) & 0.638 (0.09) & 0.721 (0.06) & 0.738 & 0.500 & 0.691 & 0.738 & \textbf{0.783} (0.02) & \textbf{0.783} (0.02) & 0.612 (0.05) \\
Breast    & 0.562 & 0.562 & 0.556 (0.00) & \textbf{0.629} (0.02) & 0.548 (0.00) & 0.594 & 0.565 & 0.623 & \textbf{0.630} & 0.550 (0.01) & 0.582 (0.02) & 0.628 (0.00) \\
Colon     & 0.548 & 0.516 & 0.500 (0.00) & 0.561 (0.09) & \textbf{0.635} (0.12) & 0.597 & 0.597 & \textbf{0.629} & 0.597 & 0.570 (0.02) & 0.594 (0.01) & 0.597 (0.00) \\
Leukemia  & \textbf{0.972} & 0.708 & 0.778 (0.00) & 0.910 (0.05) & 0.969 (0.01) & 0.931 & 0.722 & 0.861 & \textbf{0.972} & \textbf{0.972} (0.00) & 0.971 (0.01) & 0.971 (0.01) \\
Lung1     & 0.901 & 0.878 & 0.899 (0.03) & 0.832 (0.01) & 0.834 (0.14) & 0.967 & 0.967 & \textbf{0.995} & \textbf{0.995} & 0.962 (0.02) & 0.890 (0.00) & 0.989 (0.00) \\
Lung2     & 0.783 & 0.567 & 0.507 (0.00) & 0.676 (0.02) & 0.777 (0.01) & 0.783 & 0.783 & 0.724 & \textbf{0.803} & \textbf{0.788} (0.00) & 0.783 (0.00) & 0.783 (0.00) \\
Lymphoma  & \textbf{0.984} & 0.774 & 0.571 (0.02) & 0.874 (0.12) & 0.618 (0.08) & 0.935 & 0.742 & 0.968 & \textbf{0.936} & 0.853 (0.18) & 0.903 (0.02) & 0.647 (0.12) \\
Prostate  & 0.578 & 0.578 & 0.555 (0.01) & 0.568 (0.07) & 0.578 (0.01) & \textbf{0.618} & 0.588 & 0.588 & 0.569 & 0.568 (0.00) & 0.575 (0.01) & \textbf{0.616} (0.00) \\
SRBCT     & 0.556 & 0.492 & 0.543 (0.01) & 0.460 (0.04) & 0.546 (0.08) & 0.556 & 0.524 & 0.587 & \textbf{0.984} & \textbf{0.984} (0.00) & 0.981 (0.01) & 0.975 (0.02) \\
SuCancer  & 0.523 & 0.511 & \textbf{0.672} (0.00) & 0.569 (0.01) & 0.549 (0.05) & 0.667 & \textbf{0.672} & 0.500 & 0.603 & \textbf{0.687} (0.05) & 0.609 (0.02) & 0.605 (0.02) \\
\midrule
Rank      & 6.000 (3.3) & 8.800 (3.3) & 9.500 (2.9) & 8.600 (2.9) & 7.700 (3.2) & \textbf{4.100} (1.6) & 6.400 (3.6) & 5.400 (3.5) & \textbf{3.400} (2.8) & 5.000 (3.7) & 5.100 (2.2) & 4.800 (3.0) \\
Regret    & 0.109 (0.1) & 0.172 (0.1) & 0.191 (0.1) & 0.145 (0.1) & 0.139 (0.1) & 0.078 (0.1) & 0.151 (0.2) & 0.100 (0.1) & \textbf{0.041} (0.0) & \textbf{0.045} (0.0) & 0.051 (0.0) & 0.074 (0.1) \\
\bottomrule
\end{tabular}
}
\end{table*}

\begin{table*}[h]
\centering
\caption{ARI comparison of clustering methods across 10 gene microarray datasets.}
\label{ari1}
\resizebox{\textwidth}{!}{
\begin{tabular}{lcccccccccccc}
\toprule
\textbf{Dataset} & KMeans & SpecGEM & UMAP & DEC & DeepCluster & IFPCA & IFVAE & i-IF-PCA & i-IF-Lap & i-IF-Lap+UMAP & i-IF-Lap+DeepCluster & i-IF-Lap+VAE \\
\midrule
Brain     & 0.375 & \textbf{0.567} & 0.450 (0.03) & 0.411 & 0.534 (0.10) & 0.481 & 0.189 & 0.468 & 0.546 & 0.547 (0.03) & \textbf{0.552} (0.02) & 0.344 (0.05) \\
Breast    & \textbf{0.116} & 0.006 & 0.005 (0.00) & 0.010 & 0.006 (0.00) & 0.004 & 0.007 & 0.017 & \textbf{0.025} & -0.006 (0.00) & 0.013 (0.01) & 0.017 (0.00) \\
Colon     & \textbf{0.090} & -0.010 & -0.018 (0.00) & 0.030 & \textbf{0.207} (0.22) & 0.009 & 0.013 & 0.045 & 0.018 & 0.003 (0.01) & 0.016 (0.01) & 0.018 (0.00) \\
Leukemia  & 0.889 & 0.212 & 0.300 (0.00) & 0.568 & 0.642 (0.22) & 0.734 & 0.211 & 0.515 & \textbf{0.890} & \textbf{0.890} (0.00) & 0.885 (0.04) & \textbf{0.896} (0.04) \\
Lung1     & 0.487 & 0.595 & 0.464 (0.18) & 0.239 & 0.623 (0.05) & 0.834 & 0.893 & \textbf{0.973} & \textbf{0.973} & 0.830 (0.01) & 0.426 (0.01) & 0.945 (0.01) \\
Lung2     & 0.254 & -0.003 & -0.005 (0.00) & -0.002 & 0.239 (0.01) & 0.254 & 0.240 & 0.096 & \textbf{0.314} & \textbf{0.281} (0.00) & 0.254 (0.00) & 0.245 (0.00) \\
Lymphoma  & \textbf{0.947} & 0.398 & 0.402 (0.02) & 0.652 & 0.467 (0.17) & 0.824 & 0.652 & \textbf{0.893} & 0.880 & 0.738 (0.25) & 0.803 (0.02) & 0.505 (0.16) \\
Prostate  & 0.016 & 0.009 & 0.003 (0.00) & 0.009 & 0.022 (0.01) & 0.050 & 0.022 & \textbf{0.026} & 0.009 & 0.009 (0.00) & 0.013 (0.00) & \textbf{0.047} (0.00) \\
SRBCT     & 0.121 & 0.125 & 0.190 (0.01) & 0.082 & 0.162 (0.08) & 0.143 & 0.129 & 0.259 & \textbf{0.946} & \textbf{0.946} (0.00) & \textbf{0.938} (0.02) & 0.917 (0.06) \\
SuCancer  & -0.003 & 0.092 & 0.115 (0.00) & 0.011 & -0.002 (0.01) & 0.102 & \textbf{0.115} & -0.002 & 0.046 & \textbf{0.145} (0.07) & 0.045 (0.02) & 0.042 (0.02) \\
\midrule
Rank      & 5.900 (4.3) & 8.100 (3.4) & 9.300 (3.2) & 8.500 (2.5) & 6.700 (2.8) & 5.700 (3.0) & 7.200 (3.3) & 5.200 (3.3) & \textbf{3.300} (2.4) & \textbf{5.100} (3.9) & 5.500 (2.6) & 5.200 (3.3) \\
Regret    & 0.187 (0.3) & 0.317 (0.3) & 0.326 (0.2) & 0.315 (0.3) & 0.226 (0.2) & 0.173 (0.2) & 0.269 (0.3) & 0.187 (0.2) & \textbf{0.051} (0.1) & \textbf{0.078} (0.1) & 0.122 (0.2) & 0.119 (0.1) \\
\bottomrule
\end{tabular}
}
\end{table*}

\vspace{-2mm}

\section{REAL DATASETS}\label{sec:data}

We evaluate our proposed i-IF-Learn method on a collection of 18 datasets, including 10 gene microarray datasets and 8 single-cell RNA sequencing (scRNA-seq) datasets. 
The datasets are publicly available at \url{ https://data.mendeley.com/datasets/cdsz2ddv3t/1} and \url{https://data.mendeley.com/datasets/nv2x6kf5rd/1}. 
Details and pre-processing can be found in Appendix \ref{sec:app-dataset}. These datasets are characterized by high dimensionality, sparse signal structure, and varying degrees of cluster separation, making them popular in literature for assessing the performance of high-dimensional clustering algorithms.


\vspace{-1mm}
\subsection{Gene Microarray Datasets}

We benchmark our method on 10 gene microarray datasets. For each data set, we have a set of patients from different classes, and the gene expression levels of the same genes across all patients are recorded. These datasets have been widely used in prior clustering studies \citep{ifpca, chen2023subject}. 
They cover a range of cluster counts (from 2 to 5), with sample sizes ranging from 40 to 300 and number of genes typically in the thousands. Our goal is to tell the class from the gene expression levels data. 

For each dataset, we consider two variants of our method: i-IF-Lap and i-IF-PCA. 
Using i-IF-Lap as a pre-processing feature selection method, we further consider i-IF-Lap+DeepCluster, i-IF-Lap+UMAP, and i-IF-Lap+VAE, where we apply DeepCluster, UMAP, and VAE to the final influential features $\hat{I}$, respectively. 
The benchmark methods include: (1) classical methods, such as KMeans \citep{kmeans} and SpecGEM \citep{SpecGEM}; (2) neural networks, including DeepCluster \citep{caron2019deepclusteringunsupervisedlearning}, DEC \citep{DEC} and UMAP \citep{UMAP}; (3) feature selection and clustering methods, including IFPCA \citep{ifpca} and IFVAE \citep{chen2023subject}. 
Additionally, the experimental results for IDC \citep{svirsky2024interpretable} are provided in Appendix \ref{sec:appendix_idc_clear}. The compute resource is in Appendix \ref{sec:computer-resources}. Implementation details and hyperparameter selection for all algorithms are in Appendix \ref{sec:implemention}.

We assess clustering quality using two complementary metrics: \textbf{Accuracy}, which measures the best-matched proportion of true labels, and \textbf{Adjusted Rand Index (ARI)} \citep{ari}, which quantifies the similarity between predicted and true clusters for unbalanced data. When the standard deviation is smaller than $0.0001$, we do not report it. To summarize performance across datasets, we further report the average rank to rank the algorithm among all methods and the average  regret that captures how far a method’s result deviates from the best-performer on each dataset. Lower rank and regret mean better performance.

The accuracy and ARI are summarized in Table~\ref{acc1} and \ref{ari1}, respectively. In both tables, our i-IF-Lap algorithm demonstrates the most consistent and superior performance across 10 microarray datasets. 
It outperforms all other methods in 5 out of 10 datasets in terms of accuracy. In both tables, i-IF-Lap achieves the lowest average rank and regrets, indicating that it consistently ranks among the top-performing algorithms and captures all the information across diverse datasets.

Our i-IF-Lap algorithm not only suggests consistent clustering labels $\hat{\ell}$, but also recovers an influential feature set $\hat{I}$. 
Based on $\hat{I}$, DeepCluster, UMAP and VAE all enjoy lower average ranks (7.7$\to$5, 6.5$\to$5.1, and 5.4$\to$4.8, respectively), compared to their performance without i-IF-Lap pre-processing. It strongly supports the consistency of the i-IF-Lap feature selection. 

\vspace{-2mm}
\subsection{Single-cell RNA Sequencing Datasets}

We further evaluate i-IF-Learn on 8 single-cell RNA-seq (scRNA-seq) datasets, which measures gene expression levels at the resolution of individual cells. The number of cells ranges from a few hundred to several thousand, with gene dimensions ranging from 2,000 to 10,000.
Due to dropout events, scRNA-seq data are usually more sparse and noisy than gene microarray datasets. 

We consider two variants of our method, i-IF-Lap and i-IF-PCA, and further explore i-IF-Lap combined with DeepCluster, UMAP, or VAE to illustrate the effect of selected features. Benchmark methods include: (1) scRNA-seq clustering baselines, such as Seurat \citep{Seurat}, SC3 \citep{SC3}, scAMF \citep{scAMF}, and DESC \citep{DESC}; (2) neural network methods, including DeepCluster \citep{caron2019deepclusteringunsupervisedlearning} and UMAP \citep{UMAP}; and (3) feature selection combined with clustering, including IFPCA \citep{ifpca} and IFVAE \citep{chen2023subject}. Additionally, the experimental results for CLEAR \citep{CLEAR} are provided in Appendix \ref{sec:appendix_idc_clear}. Compute resources are listed in Appendix \ref{sec:computer-resources}, and implementation details with hyperparameter choices are in Appendix \ref{sec:implemention}.

\begin{table*}[h]
\centering
\caption{Accuracy comparison of clustering methods across 8 scRNA-seq datasets.}
\label{acc2}
\resizebox{\textwidth}{!}{
\begin{tabular}{lccccccccccccc}
\toprule
\textbf{Dataset} & Seurat & SC3 & scAMF & DESC & UMAP & DeepCluster & IFPCA & IFVAE & i-IF-PCA & i-IF-Lap & i-IF-Lap+UMAP & i-IF-Lap+DeepCluster & i-IF-Lap+VAE \\
\midrule
Camp1     & 0.643 & 0.788 & \textbf{0.882} & \textbf{0.799} & 0.673 (0.03) & 0.612 (0.02) & 0.738 & 0.706 & 0.738 & 0.740 & 0.687 (0.07) & 0.657 (0.03) & 0.640 (0.00) \\
Camp2     & 0.654 & \textbf{0.778} & \textbf{0.673} & 0.656 & 0.615 (0.01) & 0.608 (0.04) & 0.660 & 0.690 & 0.617 & 0.605 & 0.573 (0.00) & 0.656 (0.00) & 0.577 (0.03) \\
Darmanis  & 0.779 & 0.736 & 0.766 & 0.609 & 0.628 (0.03) & 0.583 (0.01) & 0.789 & 0.540 & 0.783 & \textbf{0.785} & 0.718 (0.04) & \textbf{0.793} (0.05) & 0.662 (0.06) \\
Deng      & 0.534 & 0.563 & 0.646 & 0.563 & 0.559 (0.09) & 0.624 (0.08) & 0.828 & 0.652 & 0.802 & \textbf{0.869} & 0.658 (0.01) & \textbf{0.857} (0.01) & 0.830 (0.07) \\
Goolam    & 0.629 & 0.758 & 0.823 & 0.629 & 0.508 (0.00) & \textbf{0.847} (0.08) & 0.721 & 0.492 & 0.629 & 0.758 & 0.500 (0.00) & 0.835 (0.09) & \textbf{0.945} (0.02) \\
Grun      & \textbf{0.993} & 0.500 & 0.523 & 0.968 & 0.663 (0.00) & 0.783 (0.02) & 0.673 & 0.750 & 0.991 & \textbf{0.994} & 0.691 (0.00) & 0.694 (0.01) & 0.724 (0.01) \\
Li        & \textbf{0.985} & 0.919 & 0.804 & 0.827 & 0.896 (0.02) & 0.871 (0.05) & 0.909 & 0.852 & \textbf{0.980} & 0.966 & 0.955 (0.00) & 0.970 (0.01) & 0.797 (0.04) \\
Patel     & 0.653 & \textbf{0.995} & \textbf{0.958} & 0.939 & 0.927 (0.01) & 0.859 (0.05) & 0.940 & 0.569 & 0.788 & 0.942 & 0.945 (0.00) & 0.954 (0.01) & 0.875 (0.05) \\
\midrule
Rank   & 7.125 (4.3) & 5.750 (4.2) & 5.625 (4.3) & 7.250 (3.4) & 9.750 (1.5) & 10.750 (3.0) & 5.625 (2.5) & 8.875 (3.1) & 5.750 (3.0) & \textbf{4.250} (3.2) & 7.750 (2.6) & \textbf{4.375} (3.1) & 8.250 (4.4) \\
Regret & 0.171 (0.2) & 0.150 (0.2) & 0.146 (0.2) & 0.156 (0.1) & 0.222 (0.1) & 0.363 (0.2) & 0.123 (0.1) & 0.249 (0.1) & 0.114 (0.1) & \textbf{0.073} (0.1) & 0.293 (0.2) & \textbf{0.103} (0.1) & 0.149 (0.1) \\
\bottomrule
\end{tabular}
}
\end{table*}

\begin{table*}[h]
\centering
\caption{ARI comparison of clustering methods across 8 scRNA-seq datasets.}
\label{ari2}
\resizebox{\textwidth}{!}{%
\begin{tabular}{lccccccccccccc}
\toprule
\textbf{Dataset} & Seurat & SC3 & scAMF & DESC & UMAP & DeepCluster & IFPCA & IFVAE & i-IF-PCA & i-IF-Lap & i-IF-Lap+UMAP & i-IF-Lap+DeepCluster & i-IF-Lap+VAE\\ 
\midrule
Camp1     & 0.519 & \textbf{0.763} & \textbf{0.801} & 0.729 & 0.521 (0.04) & 0.516 (0.02) & 0.629 & 0.639 & 0.635 & 0.650 & 0.624 (0.08) & 0.616 (0.00) & 0.597 (0.00) \\
Camp2     & 0.425 & \textbf{0.594} & 0.484 & 0.483 & 0.406 (0.01) & 0.375 (0.03) & 0.490 & 0.464 & 0.522 & \textbf{0.524} & 0.399 (0.00) & 0.505 (0.01) & 0.413 (0.04) \\
Darmanis  & \textbf{0.719} & 0.700 & 0.667 & 0.526 & 0.535 (0.05) & 0.458 (0.01) & 0.703 & 0.428 & 0.674 & 0.694 & 0.591 (0.03) & \textbf{0.703} (0.06) & 0.558 (0.06) \\
Deng      & 0.427 & 0.541 & 0.561 & 0.426 & 0.440 (0.09) & 0.393 (0.13) & 0.848 & 0.431 & 0.810 & \textbf{0.876} & 0.567 (0.01) & \textbf{0.867} (0.01) & 0.835 (0.10) \\
Goolam    & 0.544 & 0.687 & \textbf{0.914} & 0.543 & 0.423 (0.00) & 0.766 (0.01) & 0.537 & 0.205 & 0.582 & 0.687 & 0.427 (0.00) & 0.801 (0.13) & \textbf{0.976} (0.01) \\
Grun      & \textbf{0.969} & -0.060 & -0.074 & 0.928 & 0.093 (0.00) & 0.137 (0.07) & -0.096 & 0.244 & 0.955 & \textbf{0.971} & 0.145 (0.00) & 0.150 (0.01) & 0.198 (0.01) \\
Li        & \textbf{0.971} & 0.934 & 0.779 & 0.782 & 0.885 (0.02) & 0.632 (0.13) & 0.880 & 0.782 & \textbf{0.985} & 0.943 & 0.936 (0.00) & 0.948 (0.01) & 0.790 (0.04) \\
Patel     & 0.577 & \textbf{0.989} & \textbf{0.905} & 0.383 & 0.836 (0.01) & 0.729 (0.09) & 0.853 & 0.383 & 0.697 & 0.871 & 0.874 (0.00) & 0.898 (0.02) & 0.784 (0.04) \\
\midrule
Rank   & 7.250 (4.7) & 4.750 (3.5) & 6.125 (4.4) & 8.500 (3.6) & 9.750 (1.9) & 10.875 (3.3) & 7.375 (3.2) & 8.125 (3.8) & 5.375 (3.1) & \textbf{3.375} (1.8) & 7.500 (2.5) & \textbf{4.125} (2.5) & 7.250 (3.3) \\
Regret & 0.220 (0.2) & 0.221 (0.1) & 0.234 (0.3) & 0.256 (0.2) & 0.347 (0.2) & 0.363 (0.2) & 0.262 (0.3) & 0.383 (0.3) & 0.131 (0.1) & \textbf{0.087} (0.1) & 0.271 (0.2) & \textbf{0.178} (0.3) & 0.220 (0.2) \\
\bottomrule
\end{tabular}
}
\end{table*}

The accuracy and ARI are demonstrated in Tables~\ref {acc2} and \ref{ari2}, respectively. 
Our i-IF-Lap algorithm achieves the lowest average rank and average regret in both tables. The second best performer is DeepCluster with i-IF-Lap selected $\hat{I}$. It proves the power of deep clustering methods with i-IF-Lap pre-processing. As a summary, our i-IF-Lap algorithm provides both reliable clustering labels and influential feature set.

\subsection{Statistical Significance}

To rigorously evaluate performance, we conducted non-parametric statistical testing across all 18 real-world datasets. The Friedman test on the clustering results yielded $p = 3.6 \times 10^{-5}$, indicating a statistically significant difference in overall performance among the evaluated methods. 

To further assess pairwise superiority, we conducted a Wilcoxon signed-rank test with Holm correction against the alternative hypothesis that i-IF-Lap achieves better clustering. Table \ref{tab:wilcoxon} summarizes the results for methods evaluated across all 18 datasets. Our i-IF-Lap method significantly outperforms baselines such as UMAP, DeepCluster, and IFVAE, and shows statistically significant improvements over its VAE and UMAP pipeline variants.



\begin{table}[ht]
\centering
\footnotesize 
\setlength{\tabcolsep}{2.5pt} 
\caption{Summary of $p$-values from the Wilcoxon signed-rank test with Holm correction (Alternative hypothesis: i-IF-Lap works better). The methods compared are those evaluated across all 18 datasets.}
\label{tab:wilcoxon}
\begin{tabular}{lc @{\hspace{0.2mm}} lc} 
\toprule
\textbf{Method} & \textbf{$p$-value} & \textbf{Method} & \textbf{$p$-value} \\
\midrule
UMAP                 & 0.000 & i-IF-Lap+UMAP        & 0.045 \\
DeepCluster          & 0.011 & i-IF-PCA             & 0.074 \\
IFVAE                & 0.011 & IFPCA                & 0.163 \\
i-IF-Lap+VAE         & 0.045 & i-IF-Lap+DeepCluster & 0.290 \\
\bottomrule
\end{tabular}
\end{table}


\section{SYNTHETIC DATASET}\label{sec:simu}

We conduct simulation studies to evaluate i-IF-Learn under controlled settings. We first compare the feature selection and clustering performance of i-IF-Learn methods under linear and non-linear settings, and then discuss the effects of the initial label and the constant $c$ in reliability weightage. 

{\bf Linear setting}. Let $X_i \sim N(\ell_i \mu, \Sigma)$, where $\ell_i \in \{-1, 1\}$ with equal probability. 
The mean vector $\mu \in \reals^p$ has $\mu_j = 0$ for $j \notin I$ and $\mu_j \neq 0$ for $j \in I$. 
Let $I = I_s \cup I_w$, where $\mu_j \sim \frac{1}{2}N(\tau_s, 0.01^2) + \frac{1}{2}N(-\tau_s, 0.01^2)$ for $j \in I_s$ and $\mu_j \sim \frac{1}{2}N(\tau_w, 0.01^2) + \frac{1}{2}N(-\tau_w, 0.01^2)$ for $j \in I_w$. Hence, $I_s$ is the set of relatively strong signals and $I_w$ is the set of weak signals. 
The covariance matrix $\Sigma$ is a diagonal matrix with diagonals $\sigma_j^2$, where $\sigma_j \sim Unif(1, 3)$. 

We set $n = 500$, $p=5000$, $|I_s| = 4$ relatively strong signals and $|I_w| = 100$ weak signals. 
Let $\tau_s = 1.1$ and $\tau_w \in \{0.1, 0.15, 0.2, \cdots, 1.0\}$. A larger $\tau_w$ indicates a stronger signal-to-noise ratio. 

{\bf Methods.} We consider 1) i-IF-Lap and i-IF-PCA; 2) i-IF-Lap+DeepCluster/UMAP/VAE, where DeepCluster, UMAP, and VAE are applied to the influential features $\hat{I}$ from i-IF-Lap, respectively; 3) KMeans, SpecGEM, DeepCluster, DEC, UMAP, IFPCA and IFVAE as benchmark methods. 
For feature selection accuracy, we compare i-IF-Lap, i-IF-PCA, and IFPCA. For all methods, the input is the data points $X_i$'s, without labels or influential feature information. The compute resource is in Appendix \ref{sec:computer-resources}. Implementation details and hyperparameter selection for all algorithms are in Appendix \ref{sec:implemention}.

\begin{figure}[h]
    \centering
    \includegraphics[width=\linewidth]{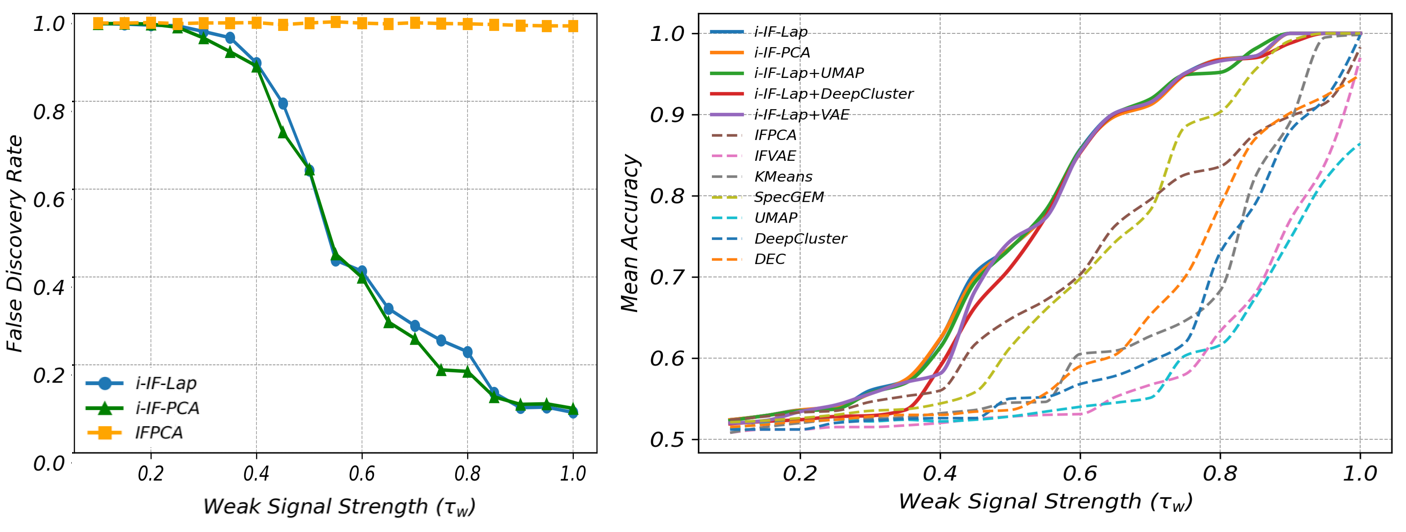}
    \caption{Left: FDR of feature selection step versus signal strength for signals in $I_w$.  Right: the clustering accuracy versus signal strength for signals in $I_w$.}
    \label{fig:comparison}
    
\end{figure}

{\bf Results.} The accuracy rates versus $\tau_w$ over 100 repetitions in the right panel of Figure \ref{fig:comparison}.  
As $\tau_w$ increases, all methods have a better accuracy from 0.5 to $\sim 1$. 
Our i-IF-Learn algorithms perform the best, especially when $0.4 \leq \tau_w \leq 0.8$. Methods using i-IF-Lap as pre-processing step also enjoy an outstanding performance, due to the reliable recovery of influential features. 

To investigate the estimate of $I$, we summarize the False Discovery Rate (FDR) over 100 repetitions in the left panel of Figure \ref{fig:comparison}. 
As $\tau_w$ increases, FDR for i-IF-Learn drops sharply, while IFPCA remains high across all settings. This highlights the benefit of iterative refinement: i-IF-Learn is able to recover weak signals with high precision. More figures about $I$ can be found in Appendix \ref{sec:appendix-synthetic}.

    


{\bf Nonlinear setting.} Sample the underlying data points from a 2D manifold, and then project them to $p$-dimensional space. The observed data points are $x_i \in \reals^p$. Let $n = 500$ and $K = 2$. We consider two experiments:

\begin{itemize}
\vspace{-1em}
    \item \textbf{p-Sweep.} 
    Let 20 features be strong signals with strength 1.0 and 60 features be weak signals with strength 0.2, while the remaining features are irrelevant. Vary $p$ from 1500 to 6000.

    \item \textbf{$\mu$-Power Sweep.} Let $p=4000$, with $80$ influential features. Each influential feature $j$ has a signal strength at $\mu_j^{a}$, where $\mu_j \sim_{i.i.d} U(0.2, 1)$ and the power $a \in \{1/4,1/3,1/2,1,2,3,4\}$.

\end{itemize}

\begin{figure}[h]
    \centering
    \includegraphics[width=\linewidth]{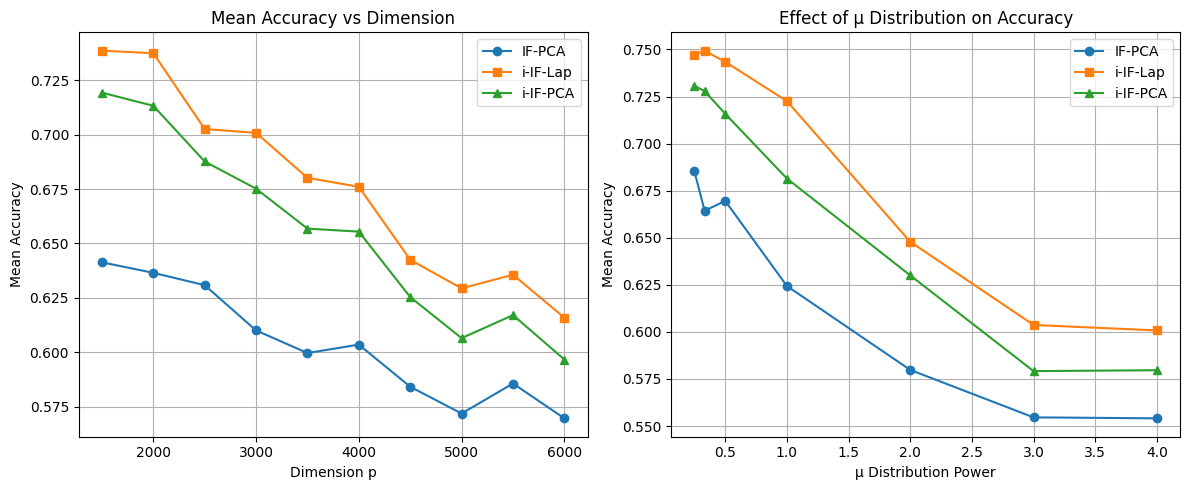}
    \caption{Left: clustering accuracy versus dimension $p$ in the $p$-Sweep experiment.  Right: clustering accuracy versus the power $a$ in the $\mu$-Sweep experiment.}
    \label{fig:combine_nonlinear}
\end{figure}

{\bf Methods}. We compare \textbf{i-IF-Lap}, \textbf{i-IF-PCA}, and \textbf{IFPCA}. 
Figure~\ref{fig:combine_nonlinear} shows that i-IF-Lap outperforms all other methods in both settings. It illustrates the power of non-linear embedding. Furthermore, the i-IF-PCA method outperforms IFPCA, indicating the power of iteration. 




{\bf Robustness.}
We examine the robustness of i-IF-Learn with respect to the choice of the constant $c$ in Eq.~\eqref{eqn:w} and the initialization scheme. 
Table~\ref{addtional_exp} reports clustering accuracies for $c \in \{0.4, 0.5, 0.6\}$ under both linear and nonlinear settings. For each $c$, the top row corresponds to IFPCA initialization, while the bottom row corresponds to a random initialization. Results comparing different embedding methods (e.g., UMAP, Autoencoder, PCA, Laplacian Eigenmaps) can be found in Appendix \ref{addtional_exp}.
The results show that i-IF-Learn is stable across different constants $c$. Furthermore, even with a random initialization, our i-IF-Learn framework still enjoys some clustering improvements.

\begin{table}[h]
\centering
\scriptsize
\setlength{\tabcolsep}{3pt} 
\setlength{\extrarowheight}{1pt}
\begin{tabular}{|c|c|cc|cc|}
\hline
\multirow{2}{*}{$c$} & \multirow{2}{*}{Init  $\ell^{(0)}$} & \multicolumn{2}{c|}{Linear} & \multicolumn{2}{c|}{Non-linear} \\
\cline{3-6}
 &  & i-IF-Lap & i-IF-PCA & i-IF-Lap & i-IF-PCA \\
\hline
\multirow{2}{*}{0.4} 
& IFPCA & 0.733 (0.22) & 0.731 (0.22) & 0.715 (0.08) & 0.674 (0.11) \\
& Random & 0.606 (0.17) & 0.611 (0.17) & 0.618 (0.11) & 0.610 (0.11) \\
\hline
\multirow{2}{*}{0.5} 
& IFPCA & 0.713 (0.22) & 0.712 (0.22) & 0.701 (0.09) & 0.674 (0.10) \\
& Random & 0.576 (0.12) & 0.573 (0.12) & 0.627 (0.11) & 0.611 (0.09) \\
\hline
\multirow{2}{*}{0.6} 
& IFPCA & 0.742 (0.22) & 0.738 (0.22) & 0.702 (0.09) & 0.672 (0.12) \\
& Random & 0.585 (0.14) & 0.580 (0.14) & 0.644 (0.11) & 0.619 (0.11) \\
\hline
\end{tabular}
\caption{Accuracy (mean (std)) of i-IF-Lap and i-IF-PCA across different constants and initializations.}
\label{addtional_exp}
\end{table}

\vspace{-3mm}
\section{DISCUSSION}
We introduce i-IF-Learn, an iterative framework for high-dimensional clustering that integrates feature selection with low-dimensional embedding. By adaptively using supervised pseudo-labels and unsupervised statistics, our novel screening metric enables robust feature selection even when early clustering assignments are noisy. Unlike static pipelines, i-IF-Learn iteratively refines both feature sets and labels to effectively amplify weak signals. Beyond assigning cluster labels, i-IF-Learn outputs an interpretable set of influential features.



As an exploration of the framework's flexibility, we replaced the IF step with a supervised Lasso penalty; however, this yielded sub-optimal performance (see Appendix \ref{sec:appendix_lasso}).

While demonstrating strong empirical results, i-IF-Learn presents several avenues for future research. First, the current marginal screening step evaluates features individually, potentially ignoring pairwise or block interactions. Future work could incorporate a supervised recovery step (e.g., using CIFE \citep{CIFE}, or JMI \citep{JMI}) on the full feature set, guided by the generated pseudo-labels, to capture joint effects and eliminate redundancies. Second, i-IF-Learn currently identifies a single global set of influential features. Developing high-resolution methods to detect distinct, cluster-specific feature subsets is a highly relevant next step. More HDLSS datasets can be found in the publicly available repository provided by \cite{li2017feature} (\url{https://jundongl.github.io/scikit-feature/datasets}). The implementation of i-IF-Learn is publicly available at
\url{https://github.com/mc25800852/i_if_learn}.

\subsubsection*{Acknowledgements}

This research was supported by the Singapore Ministry of Education Academic Research Fund Tier 1 under Grant A-8001451-00-00. C. Ma gratefully acknowledges the financial support from the Southern University of Science and Technology (SUSTech) during the exchange program at the National University of Singapore (NUS), and the travel support provided by the Wallinska resestipendiet and the Swedish Association for Medical Statistics (FMS) travel scholarship. The authors also thank the anonymous reviewers for their valuable comments and suggestions, which led to the addition of new experiments in this work.

\bibliography{sample}

\section*{Checklist}

\begin{enumerate}
\item For all models and algorithms presented, check if you include:
\begin{enumerate}
\item A clear description of the mathematical setting, assumptions, algorithm, and/or model. [Yes] We provide the mathematical setting and assumptions in Section \ref{sec:model}, and describe the algorithms in Section \ref{sec:algo} and provide pseudo-codes for each algorithms in Appendix \ref{sec:app-pseudocode}.
\item An analysis of the properties and complexity (time, space, sample size) of any algorithm. [Yes] The computational complexity is analyzed in Section \ref{sec:algo}.
\item (Optional) Anonymized source code, with specification of all dependencies, including external libraries. [Yes] An anonymized implementation with dependencies is included in the supplemental file \texttt{i-IF-Lear.zip} and \texttt{NumericalStudy.zip}.
\end{enumerate}

\item For any theoretical claim, check if you include:
\begin{enumerate}
\item Statements of the full set of assumptions of all theoretical results. [Yes] All assumptions are clearly stated in Section \ref{sec:model}.
\item Complete proofs of all theoretical results. [Yes] Full proofs are provided in Appendix \ref{sec:proof}.
\item Clear explanations of any assumptions. [Yes] Explanations are given alongside the assumptions in Section \ref{sec:model}.
\end{enumerate}

\item For all figures and tables that present empirical results, check if you include:
\begin{enumerate}
\item The code, data, and instructions needed to reproduce the main experimental results (either in the supplemental material or as a URL). [Yes] Code is included in the supplemental material, and datasets are public with download links provided in Section \ref{sec:data} and more details about in \ref{sec:app-dataset}.
\item All the training details (e.g., data splits, hyperparameters, how they were chosen). [Yes] Detailed implementation settings and hyperparameters are listed in Appendix \ref{sec:implemention}.
\item A clear definition of the specific measure or statistics and error bars (e.g., with respect to the random seed after running experiments multiple times). [Yes] We report standard deviations for all results, and omit them only when the variance is negligible (less than 0.0001).
\item A description of the computing infrastructure used. (e.g., type of GPUs, internal cluster, or cloud provider). [Yes] As described in Appendix \ref{sec:computer-resources}, our experiments are lightweight and reproducible on a standard PC without specialized hardware.
\end{enumerate}

\item If you are using existing assets (e.g., code, data, models) or curating/releasing new assets, check if you include:
\begin{enumerate}
\item Citations of the creator If your work uses existing assets. [Yes] All datasets and existing codes are properly cited (see Section \ref{sec:data} and Appendix \ref{sec:app-dataset}).
\item The license information of the assets, if applicable. [Yes] License information for the datasets is listed in Appendix.
\item New assets either in the supplemental material or as a URL, if applicable. [Yes] We provide an anonymized implementation of our proposed method in \texttt{i-IF-Learn.zip}.
\item Information about consent from data providers/curators. [Not Applicable] All datasets used are publicly available.
\item Discussion of sensible content if applicable, e.g., personally identifiable information or offensive content. [Not Applicable] The datasets do not contain sensitive or personally identifiable content.
\end{enumerate}

\item If you used crowdsourcing or conducted research with human subjects, check if you include:
\begin{enumerate}
\item The full text of instructions given to participants and screenshots. [Not Applicable] Our work does not involve crowdsourcing or human subjects.
\item Descriptions of potential participant risks, with links to Institutional Review Board (IRB) approvals if applicable. [Not Applicable] Our work does not involve crowdsourcing or human subjects.
\item The estimated hourly wage paid to participants and the total amount spent on participant compensation. [Not Applicable] Our work does not involve crowdsourcing or human subjects.
\end{enumerate}
\end{enumerate}

\clearpage
\appendix
\thispagestyle{empty}

%
%





%

%

\onecolumn

\onecolumn
\aistatstitle{i-IF-Learn: Iterative Feature Selection and Unsupervised Learning for High-Dimensional Complex Data: Supplement Material}
\section{TTECHNICAL DETAILS AND PROOFS}\label{sec:proof}
In section \ref{sec:weight}, we explain the details of the weight selection. With the weight selection, we prove the main theorems and corresponding lemmas in the following subsections. 

\subsection{Dynamic Weigh}\label{sec:weight}
To decide the reliability constant $w^{(t)}$, we want to test our trust in $P_{F,j}^{(t)}, j \in I^{(t-1)}$. In other words, we wonder whether these statistics follow a null distribution or not. 

Let $\pi_I^{(t-1)} = \{P_{F,j}^{(t)}, j \in I^{(t-1)}\}$ be the set of $p$-values from corrected marginal $F$-statistics, restricted on the features in $I^{(t-1)}$. 
An unreliable feature set $I^{(t-1)}$ will have uniformly distributed $p$-values $\pi_I^{(t-1)}$. 
Therefore, it is the hypothesis testing problem that $\pi_I^{(t-1)} \sim Unif(0, 1)$. 

We construct a test for the hypothesis 
\[
H_0: \pi_{(j)}^{(t - 1)} \sim Unif(0, 1) \quad 
vs
\quad 
H_1: \pi_{(j)}^{(t - 1)} \sim (1 - \epsilon) Unif(0, 1) + \epsilon G,
\]
where $G$ is some other distribution that focuses on small $p$-values. 

We conduct the Higher Criticism statistic \cite{HCT_intro} for this test. Denote $s^{(t-1)} = |I^{(t-1)}|$ and $\pi_{(j)}^{(t-1)}$ as the $j$-th smallest value in $\pi_I^{(t-1)}$. 
The $p$-value is  
\begin{equation}\label{eqn:w1}
    p_1^{(t)} = 1 - \exp(-e^{c - b T}), \mbox{ where }
    T = \max_{1 \leq j \leq 2s^{(t-1)}/3} \sqrt{s^{(t-1)}} \frac{j/s^{(t-1)}-\pi_{(j)}^{(t-1)}}{\sqrt{\pi_{(j)}^{(t-1)}(1-\pi_{(j)}^{(t-1)})}}, 
\end{equation}
and $b=\sqrt{2\log(\log(s^{(t-1)}))}$,  $c=2\log(\log(s^{(t-1)}))+\log(\log(\log(s^{(t-1)})))/2-\log(4\pi)/2$. 
Here, the $p$-value $p_1^{(t)} \in (0, 1)$, and a smaller $p_1^{(t)}$ indicates a larger possibility that $I^{(t-1)}$ is informative. 

\subsection{Proof of Theorem \ref{thm:fs}}
Given a label $\hat{\ell}$, we use it as a pseudo-label to select the influential features for the next step. The selection is based on our new score $S_j^{(t)} = S_j(\hat{\ell})$, defined in \eqref{eqn:s}, where the weight is defined in \eqref{eqn:w}. 
Therefore, to show that our IF step is powerful, we need the score $S_j(\hat{\ell})$ to efficiently evaluate the dependency between feature $x_j$ and $\hat{\ell}$, and then the data-driven threshold HCT to select a proper threshold. 

The following lemma explains the power of our statistic: when the initial label delivers some information, then the weight will be close to 1, so that the $S_j(\hat{\ell})$ highly depends on the $F$-statistics. Further, even with noisy labels, our statistic $S_j(\hat{\ell})$ clearly separates $I$ and $I^c$. 
The proof of the lemma can be found in Section \ref{subsec:lemfs}.
\begin{lemma}\label{lem:fs}
Fix a constant $q>4$. 
Denote $w_{ij} = E[\Sigma^{-1/2}X_{ij}]$ as the expectation for data point $i$ on feature $j$, and the overall mean $\bar{w}_j=\frac{1}{n}\sum_{i=1}^n w_{ij}$. Denote $n_k(\hat{\ell}) = \sum_{i=1}^n 1\{\hat{\ell}_i = k\}$, $k \in [K]$. Suppose for the influential features in $I$, the community label satisfies that for a constant $c_0 > 0$, 
\[
\min_{j\in I} U_j\geq  c_0(\log p)^2, 
\quad \text{where }
U_{j}:=\sum\nolimits_{k\in[K]}\frac{1}{n_k(\hat{\ell})} \sum\nolimits_{\hat{\ell}_i = k}(w_{ij} -\bar{w}_j)^2. 
\]
Then we have $P(w\geq 1-p^{-q/2})\geq 1-p^{-q/2}$, and further
\begin{itemize}
    \item If $j\in I$, $P(S_j(\hat{\ell})>p^{-q})\leq p^{-q}$. 
    \item If $j\in I^c$, for any $u\in (0,1)$, $P(S_j(\hat{\ell})<u)\leq u-\exp(-p)$. 
\end{itemize}
\end{lemma}

For notation simplicity, we write $S_j = S_j(\hat{\ell})$ when there is no misunderstanding. The proof consists of three steps. We first show that $S_j$ for $j \in I$ and $j \notin I$ has a clear division with high probability. Then we show that Higher Criticism Threshold (HCT) almost achieves the optimal division, in the sense that $HC(j)$ for all $j \in I^c$ is smaller than the selected HCT and no more than $C\log p$ feature in $I$ have lower $HC(j)$ than HCT. 

\textbf{Step 1: Clear division with high probability.}
We define a good event that the statistics from $j \in I$ and $j \notin I$ are clearly divided, that 
\[
B=\{\max_{j\notin I^c}S_j< 1+q\log p <\min_{j\in I}S_j\}.
\]
By Lemma \ref{lem:fs}, the probability of $B$ bounded by the union bound 
\begin{eqnarray*}
1-P(B) &\leq &|S| p^{-q}+p p^{-q}=o(1).
\end{eqnarray*}

\textbf{Step 2: Under $B$, all indices from $I$ will be selected by HCT.}
By the algorithm, $\hat{I}$ is achieved by selecting all the features with $p$-values smaller than the threshold $\tau$. 
In other words, if we order the features in the way that the p-value of $S_j$ is decreasing, then the first $|\hat{I}|$ features are selected. 

According to the definition of $B$ in Step 1, features from $I$ always have smaller $p$-values than those in $I^c$. 
Therefore, the first $|I|$ features are from $I$ and the following features are from $I^c$. As long as we set the threshold as the p-value of the $|I|$-th smallest one, then we exactly recover $I$. 
Note that we select $|\hat{I}|$ according to where the HC scores $HC(j)$ achieves maximum. If $HC(j)$ achieves maximum around $j = |I|$, then the cutoff is correct. 
It suffices to show that for any $j < |I|$, 
\begin{equation}
\label{tmp:step2}
HC(j)\leq HC(|I|).  
\end{equation}
Under $B$, $\max_{j\in I} S_j\leq p^{-q}$. Introduce it into $HC(|I|)$, we have
\begin{equation}\label{eqn:HCSlower}
HC(|I|)\geq \frac{|I|/p-p^{-q}}{\sqrt{|I|/p(1-|I|/p)}}\geq 
\frac{|I|/p-p^{-q}}{\sqrt{|I|/p(1-|I|/p)}}.
\end{equation}
Meanwhile, we have $HC(j)\leq \frac{\sqrt{(|I|-1)/p}}{\sqrt{1-(|I|-1)/p}}$ for all $j\leq |I|-1$. 

Plug these upper bounds and \eqref{eqn:HCSlower} into \eqref{tmp:step2}. We have
\begin{align*}
[HC(j)]^2 & \leq  [HC(|I|)]^2\Leftrightarrow\\
\frac{(|I|-1)/p}{1-(|I|-1)/p} & \leq \frac{(|I|/p-p^{-q})^2}{|I|/p(1-|I|/p)}\Leftrightarrow\\
(|I| - 1)|I|(p - |I|) & \leq  (|I| - p^{-q})^2 (p - |I|+1).
\end{align*}
Rearrange the terms and it suffices to show \eqref{tmp:step2} if we can show that 
\[
|I|(p - |I|)(1 - 2p^{-2q}) + |I|(|I| - 2p^{-q}) + p^{-2q}(p - |I|+1)\geq 0.
\]
When $p$ is sufficiently large, it holds. So \eqref{tmp:step2} holds for sufficiently large $p$.

\textbf{Step 3: Under $B$, only $C\log^2 p$ features from $I^c$ will be selected by HCT.} According to previous analysis, it suffices to show $HC(|I|+k)<HC(|I|)$ when $k>C\log^2 p$. 

Consider a sequence of threshold $v_k = k/|I^c|$ for $k > C_1\log^2 p$. 
Note that $P(S_j\leq u)\leq u-\exp(-p)$ for all $j \in I^c$. By the Bernstein inequality with $\delta_k = 4\sqrt{{v_k\log p}/{p}}$, when $C_1 > 2(p/|I^c|)^2$ and $|I^c|/p > 1/2$, we have 
\begin{eqnarray}
P\bigl(\bigl|\frac{1}{|I^c|}\sum_{j\in I^c} 1_{S_j<v_k}-v_k\bigr| > \delta_k\bigr) & \leq & \exp\left(-\frac{|I^c|\delta_k^2}{2(v_k(1-v_k)+\delta_k /3)}\right)\nonumber\\
& \leq & \exp\left(-\frac{|I^c|\delta_k^2}{4v_k(1-v_k)}\right) + e^{-{3|I^c|\delta_k}/{4}}\nonumber\\
& \leq & 2p^{-2}.
\end{eqnarray}
Naturally, the union bound follows, which is 
\[
P\bigl(\bigl|\frac{1}{|I^c|}\sum_{j\in I^c} 1_{S_j<v_k}- v_k\bigr|>\delta_k, \text{ for any $k \geq C_1\log^2 p$}\bigr)\leq \frac{1}{p}. 
\]
Hence, the complementary event happens with probability $1 - O(p^{-1})$. Define it as 
\[
C:=\left\{\bigl|\frac{1}{|I^c|}\sum_{j\in I^c} 1_{S_j<v_k}- v_k\bigr|<\delta_k, \text{ for any $k \geq C_1\log^2 p$}\right\}. 
\]
Denote the ordered $S_i$ as $S_{(1)}\leq S_{(2)}\leq\ldots\leq S_{(p)}$. Under the event $B$,  $S_{(|I|+k)} = S^0_{(k)}$ where $S^0_{(\cdot)}$ are the ordered $p$-values of $i \in I^c$. Now we want to derive a lower bound of $S_{(k)}$. 

Under $C$, recall that $v_j = j/|I^c|$ and $\delta_j = 4\sqrt{v_j \log p/p}$, when $j > C_1\log^2p$, 
\[
\sum_{i \in I^c} 1\{S_i < v_j\} < |I^c|(v_j + \delta_j) < j + 4\sqrt{j \log p}.
\]
It means $S^0_{(j + 4\sqrt{j \log p})} \geq v_j$. 
Now let $k$ be the smallest integer so that $k \geq j + 4\sqrt{j\log p}$, then it follows $j \geq k-1 - 4\sqrt{k \log p}$. 
Hence, under $C\cap B$, for $k \geq 3C_1\log^2p$,
\[
S_{(|I|+j + 4\sqrt{j \log p})} \geq v_j\Longrightarrow  S_{(|I|+k)} \geq (k-1)/p - 4\sqrt{k \log p}/p,
\]
This further leads to 
\begin{align*}
HC(|I|+k)=\frac{\frac{|I|+k}{p}-S_{(|I|+k)}}{\sqrt{\frac{(|I|+k)}{p}(1-\frac{|I|+k}{p})}}\leq \frac{\frac{|I|+k}{p}-\frac{k-1}{p}+\frac{4\sqrt{k\log p}}{p}}{\sqrt{\frac{(|I|+k)}{p}(1-\frac{|I|+k}{p})}}.
\end{align*}

Combine it with \eqref{eqn:HCSlower}. To conclude our claim, we need to show 
that if $0.25p>k\geq 3C_1\log^2 p$, there is  
\begin{align*}
 \frac{(|I|-p^{-c})^2}{{|I|(p-|I|)}}>    \frac{(|I|+1+4\sqrt{k\log p})^2}{(|I|+k)(p-|I|-k)}.
\end{align*}
This can be shown if 
\[
(|I|-p^{-c})^2(|I|+k)(p-|I|-k)>(|I|+1+4\sqrt{k\log p})^2(|I|+k)(p-|I|-k).
\]
When $|I| \gg \log p$, it can further be simplified as 
\[
p|I|^3+kp|I|^2>p|I|^3+8p\sqrt{k\log p}|I|^2+\text{lower order terms}. 
\]
It holds when $k\geq 3C_1\log^2 p$ and $p$ large enough. In other words, $HC(|I|+k)<HC(|I|)$ for $k > C\log^2 p$ where $C = 3C_1$. This gives our second bound. 

Combining the results, Theorem \ref{thm:fs} is proved.

\subsection{Proof of Lemma \ref{lem:fs}}\label{subsec:lemfs}
In this section, we prove Lemma \ref{lem:fs}used in the previous section. 
First we discuss the required condition that $\min_{j \in I} U_j \geq c_0(\log p)^2$, which is about the estimated labels $\hat{\ell}$ and the influential features. 

Consider the $K=2$ case where both classes have size $n/2$. For feature $j \in I$, since the overall sum is $m = 0$, for a constant signal strength $\kappa$, there is $m_1 = \kappa$ and $m_2 = - \kappa$. 
Now, consider the estimated label $\hat{\ell}$. Suppose it classifies $r_{11} n/2$ with true label 1 and $r_{21} n/2$ with true label 2 as Class 1. 
Then we have
\begin{align*}
\sum_{k\in[K]}\frac{1}{n_k}(\sum_{i\in[n_k]}m_{i,k})^2=&\frac{1}{(r_{11} + r_{21}) n/2}(\kappa r_{11}  n/2 -
 \kappa r_{21} n/2)^2\\
&+
\frac{1}{(2-r_{11} - r_{21})n/2}(\kappa (1  - r_{11}) n/2 -
 \kappa (1-r_{21}) n/2)^2\\
&= \frac{n \kappa^2 (r_{11} - r_{21})^2}{(r_{11} + r_{21})(2 - r_{11} - r_{21})}. 
\end{align*}
Hence, as long as $r_{11} \neq r_{21}$, i.e. there is a constant difference between the proportion the two classes that $\hat{\ell}$ classifies into class 1, then the whole term is at an order of $Cn\kappa^2$. The condition follows that 
\[
C n\kappa^2 \geq (\log p)^2 \Leftrightarrow 
\kappa^2 \geq \frac{(\log p)^2}{n}.
\]
It means a constant error rate is accepted in the initial label. 

Now we come to the proof. 
We will suppress the appearance of feature index $j$ in the subscript for notational simplicity. We also remove $\hat{\ell}$.  Denote $J_k = \{i \mid \hat{\ell} = k\}$, $k \in [K]$. Let $n_k = |J_k|$ for $k \in [K]$.

When $j\notin I^c$,  the nominator of the F-statistics is given by 
\[
N_{0}=\frac{1}{K-1}\sum_{k\in [K]}n_k(\frac{1}{n_k}\sum_{i\in J_k}\xi_{i}-\frac{1}{n}\sum_{i\in[n]}\xi_{i})^2\sim \frac{\chi^2_{K-1}}{K-1},
\]
where $\xi_{i}$ are some $N(0,1)$ distributed random variables. While the denominator 
\[
D_{0}=\frac{1}{n-K}\sum_{k\in [K]}\sum_{i\in J_k}(\xi_{i}-\frac{1}{n_k}\sum_{j\in J_k}\xi_{j})^2\sim \frac{\chi_{n-K}^2}{n-K}. 
\]
Using concentration inequality, we find that there is a constant $C_0$ so that 
\[
P(N_0>1+C_0 q\log p )\leq p^{-q},\quad 
P(D_0<1-C_0\sqrt{q\log p/n})\leq p^{-q}.
\]
So 
\[
P(N_0/D_0>1+2C_0 q\log p )\leq 2p^{-q}. 
\]
And if $j\in I$, denote $\Delta m_i=m_i-\bar{m},\Delta \xi_i=\xi_i-\bar{\xi}$
\begin{align*}
N_{1}=\frac{1}{K-1}\sum_{k\in [K]}n_k(\frac{1}{n_k}\sum_{i\in J_k}(\Delta\xi_{i}+\Delta m_{i}))^2\sim \frac{\chi^2_{K-1}}{K-1}+Q+U_j
\end{align*}
Where 
\[
Q=\frac{1}{K-1}\sum_{k\in[K]}\frac{1}{n_k}\left(\sum_{i\in J_k}\Delta \xi_i+\Delta m_i\right)^2
\]
Using concentration inequality of Gaussian, $P(|Q|+\frac{\chi^2_{K-1}}{K-1}>q\log p)\leq p^{-q}$. So $P(N_1<U_j-q\log p)\leq p^{-q}$.

Meanwhile, if $l(i)=k$, we denote $\nabla m_i=m_i-\frac{1}{n_k}\sum_{j\in J_k}m_j,\nabla \xi_i=\xi_i-\frac{1}{n_k}\sum_{j\in J_k}\xi_j$
\[
D_{1}=\frac{1}{n-K}\sum_{i\in [n]}(\nabla m_i+\nabla \xi_i)^2\sim 
\frac{\chi^2_{n-K}+2\sum_{i\in [n]} \nabla m_i \nabla \xi_i+(\nabla m_i)^2}{n-K}
\]
Using Gaussian concentration, there is a constant $C$, so that $P(D_1>C)\leq p^{-q}$. In combine, we find $P(D_1/N_1\leq q\log p)\leq P(D_1/N_1<U_j/C)\leq p^{-q}$. 
Therefore, 
\[
P(P_{F,j}\geq p^{-q})\leq P(D_1/N_1\leq q\log p)\leq p^{-q}. 
\]

The next step would be considering the reliability constant. We note that 
\[
T\geq \sqrt{|I|}\frac{1/|I|-P_{F,(1)}}{\sqrt{P_{F,(1)}}}\geq \frac{p^{-1}-p^{-q}}{p^{-q/2}}\geq p^{q/2-1}\geq 2p
\]
Then 
\[
p_1\geq 1- \exp(-\exp(\log p-p))\geq \exp(-p). 
\]
And we get $w=1-\frac{p_1}{p_1+0.6}\geq 1-\exp(-p)$. 

Finally, for $j\in I$ we have 
\[
P(S_j>2p^{-q})\leq P(P_{F,j}>2p^{-q}-w+1)\leq p^{-q}. 
\]
For $j\in I^c$, we have 
\[
P(S_j<u)\leq P(P_{F,j}w<u)\leq u-\exp(-p). 
\]
The result is proved. 

\subsection{Proof of Theorem \ref{thm:clustering}}

With a correct selection of $I$, $k$-means on the low-dimensional embeddings will give a nice clustering result. 
The clustering error rate is evaluated by the Hamming error, which is the proportion of unmatched labels under the best scenario. 
In detail, for estimated label $\hat{\ell}$, let $\pi:[K] \to [K]$ be any permutation of $\{1,\cdots, K\}$, then 
\[
Err(\hat{\ell}, \ell) = \min_{\pi: [K] \to [K]}\sum_{i = 1}^n 1\{\hat{\ell}_{\pi(i)} \neq \ell_i\} .
\]
For notation simplicity, denote $X$ as the post-selection data matrix $X^{(t)}$. 
Denote $s_I$ be the number of informative features in $I^{(t)}$ and $s$ be the total number. 

The normalized data matrix can be written as $W = LM_I + E$, where $E$ is the noise matrix, $L \in \{0, 1\}^{n \times K}$ is the label matrix and $M_I \in \mathbb{R}^{s \times K}$ be the mean matrix on $I^{(t)}$ among all classes. Denote $\tau_I$ be the eigengap of $M_I$, which is no smaller than $\tau \sqrt{s_I}$.  

According to random matrix theory \cite{rmt}, with high probability 
\[
\|E\|\leq 2(\sqrt{n}+\sqrt{s})
\]
Let $\hat{U}$ denote the top $K$ left singular vectors of $X$ and $U$ denote the top $K$ left singular vectors of $L M_I$. 
By Davis-Kahan Theorem \cite{daviskahan}, there exists an orthogonal matrix $O$, so that 
\[
\|\hat{U}-UO\|\leq \frac{\sqrt{n}+\sqrt{s}}{\text{eigengap}(LM_I)} \leq C \frac{\sqrt{n}+\sqrt{s}}{\sqrt{n}\tau_I}:=\delta. 
\]

Next we examine the performance of $k$-means on $\hat{U}$. For any estimated label $\hat{\ell}$ and centers $\hat{u}$, define the within-cluster distance 
$L(\hat{\ell}, \hat{u})=\sum_{i=1}^n \| z_i-\hat{u}_{\hat{\ell}(i)}\|^2$.
The algorithm $k$-means is to find $\hat{\ell}$ that minimizes $L(\hat{\ell}, \hat{u})$. 

Suppose the singular value decomposition $LM_I = U\Lambda V'$, then $U=LM_IV\Lambda^{-1}$. 
For two nodes $i$ and $j$ in the same group, there is $\ell_i = \ell_j$ and the $i$-th row and $j$-th row in $L$ are the same. Therefore, the $i$-th row and $j$-th row of $U$ are the same. 
This will be the basis of clustering. 

Denote the $i$-th row of $\hat{U}$ as $z_i$. 
Denote the $k$-th row of $M_IV\Lambda^{-1}$ as $u_k$. 
For the true labels $\ell$ and centers $u_k$'s, there is $L(\ell, u)=\sum_{i=1}^n\| z_i -u_{\ell(i)}\|^2$. 
Let $\hat{\ell}$ and $\hat{u}$ be the labels and centers identified by $k$-means. Hence, for centers $\hat{u}_k$ and labels $\hat{\ell}$, 
\begin{align}
\label{tmp:Lupp}
L(\hat{\ell}, \hat{u})\leq L(\ell, u),
\end{align}
For any community $k$, let the permutation $\pi(k) = \arg\min_{1 \leq j \leq K} \|u_k - \hat{u}_{j}\|$. Hence, $\pi(k)$ is the community where the estimated center is closest to $u_k$.  We want to control the distance between $u_k$ and $\hat{u}_{\pi(k)}$. According to $k$-means, for community $k$, let $n_k$ be the size of community $k$, then 
\begin{align}
\label{tmp:Llower}
L(\hat{\ell}, \hat{u})&\geq \sum_{\ell(i)=k} \| z_i -\hat{u}_{\hat{\ell}(i)}\|^2\nonumber\\
&\geq  \sum_{\ell(i)=k} (-\| z_i-u_k O\|^2+\frac12\|u_k O-\hat{u}_{\hat{\ell}(i)}\|^2)\nonumber\\
&\geq  -\sum_{i=1}^n\| z_i-u_kO\|^2+\frac12\sum_{\ell(i)=k} \|u_k O-\hat{u}_{\hat{\ell}(i)}\|^2\nonumber\\
\quad &\geq  - L(\ell, u) +\frac12 n_k\|u_k O-\hat{u}_{\pi(k)}\|^2.
\end{align}
Combining \eqref{tmp:Lupp} and \eqref{tmp:Llower}, there is 
\[
\|u_k-\hat{u}_{\pi(k)}\|^2 \leq 4L(\ell, u)/n_k \leq 4L(\ell, u)/c n.
\]
Recall that the centers $\{u_i, i=1,\ldots, K\}$ are $1/\sqrt{n}$-distance apart.
If $L(\ell, u) < c_0$ for a constant $c_0$ small enough, then each $u_k$ is paired with a unique $\hat{u}_{\pi(k)}$ such that $\|\hat{u}_{\pi(k)}-u_k\|\leq 2\sqrt{L(\ell, u)/c n}$. 

Furthermore, since the data points come from $\hat{U}$ and the centers are from $U$, where the distance is controlled by the Davis-Kahan Theorem. 
Therefore, we control the loss in the idean case, where 
\begin{align}
\label{tmp:Lcontrol}
L(\ell, uO)&=\sum_{i=1}^n\| z_i O'-u_{\ell(i)} OO'\|^2\leq K\delta^2.
\end{align}

Finally, we consider the mis-classification rate. To simplify the notations, we assume $\pi(k) = k$ without loss of generality. Then the misclassified nodes are $S = \{i: \ell_i \neq \hat{\ell}_i\}$. The misclassification rate is $Err(\hat{\ell}, \ell) = |S|/n$. 
\begin{align*}
 L(\hat{\ell},\hat{u})=\sum_{i=1}^n\| z_i-\hat{u}_{\hat{\ell}_i}\|^2&\geq -\sum_{i=1}^n\| z_i O'-u_{\ell_i}\|^2+\frac12\sum_{i=1}^n\|u_{\ell_i}-\hat{u}_{\hat{\ell}_i}O'\|^2\\
&\geq  - L(\ell, uO) +\frac12\sum_{i \in S} \|u_{\ell_i}-\hat{u}_{\hat{\ell}_i}O'\|^2\\
&\geq - L(\ell, uO) +\frac12 |S|/n.
\end{align*}
Combining it with $L(\hat{\ell},\hat{u}) \leq L(\ell, uO)$ in (\ref{tmp:Lupp}) and $L(\ell, uO) \leq  \delta^2$ in (\ref{tmp:Lcontrol}), then we have $Err(\hat{\ell}, \ell) = |S|/n \leq n \delta^2$. 
 The theorem is proved.
\newpage
\section{PSEUDO-CODE FOR ALGORITHMS}\label{sec:app-pseudocode}
In this section, we present the pseudo-code for our algorithm and some other algorithms without existing packages. 

Here is the list of algorithms we have discussed: 
\begin{itemize}
    \item IFPCA, the initialization step and a comparison algorithm, in Algorithm \ref{alg:ifpca}
    \item i-IF-Learn, our algorithm, in Algorithm \ref{alg:iiflearn}
    \item DeepCluster in Algorithm \ref{alg:deepcluster}
    \item Deep Embedding Clustering (DEC) in Algorithm \ref{alg:dec}
    \item Uniform Manifold Approximation and Projection (UMAP) in Algorithm \ref{alg:umap}
    \item Variational Autoencoder (VAE) in Algorithm  \ref{alg:vae}

\end{itemize}

In this section, we only present the pseudo-code. Hyper-parameter selections and implementation details can be found in Section \ref{sec:implemention}.
\begin{algorithm}[H]
\caption{IFPCA Initialization Procedure}
\label{alg:ifpca}
\begin{algorithmic}[1]
\Require Data matrix $X \in \reals^{n \times p}$, number of clusters $K$
\Ensure Initial cluster labels $\ell^{(0)}$, initial influential feature set $I^{(0)}$, p-values of KS test $ P_{KS,p}$
    \State Normalized data matrix $X$, denoted it as $W$.
    \Statex \textbf{Step 1.1 Compute Kolmogorove-Smirnov scores}
            \For{$j = 1$ to $p$}
                \State $\psi_{n,j} \gets \sqrt{n} \cdot \sup_{t} \left| F_{n,j}(t) - \Phi(t) \right|$, where $F_{n,j}(t)$ is the empirical cumulative density function of $w_j$ and $\Phi$ is standard normal distribution. 
            \EndFor
            \State Normalize scores: $\psi^*_{n,j} \gets \dfrac{\psi_{n,j} - \text{mean}(\psi_{n,\cdot})}{\text{std}(\psi_{n,\cdot})}$
            
    \Statex \textbf{Step 1.2: HCT and feature selection:}
    \For{$j = 1$ to $p$}
        \State $P_{KS,j} \gets 1 - F_0(\psi^*_{n,j})$, where $F_0$ is the null distribution.
    \EndFor
    \State Sort p-values: $P_{KS,1} \leq P_{KS,2} \leq \dots \leq P_{KS,p}$
    
    \For{$j = 1$ to $p/2$ \textbf{where} $P_{KS,j} > \log(p)/p$}
        \State $HC_{p,j} \gets \dfrac{\sqrt{p} (j/p - \pi_{(j)})}{\sqrt{\max\{ \sqrt{n} (j/p - \pi_{(j)}), 0 \}} + j/p}$
    \EndFor
    \State $\hat{j} \gets \arg\max_j HC_{p,j}$, $t_p^{\text{HC}} \gets \psi^*_{n, \hat{j}}$
    \State $I^{(0)} \gets \{1\leq j \leq p \mid \psi^*_{n,j} > t_p^{\text{HC}} \}$
    
    \vspace{1mm}
    \State \textbf{Step 1.3: PCA embedding and $k$-means clustering:}
    \State Apply PCA to post-selection data, retain top $K-1$ components. Denote it as $U$.
    \State $\ell^{(0)} \gets$ $k$-means({$U$, $K$})
\end{algorithmic}
\end{algorithm}

\begin{algorithm}[H]
\caption{i-IF-Learn}
\label{alg:iiflearn}

\begin{algorithmic}[1]
\Require Data matrix $X \in \reals^{n \times p}$, number of clusters $K$
\Ensure Predicted cluster labels $\ell$, influential feature set $I$

\vspace{1mm}
\Statex \textbf{Step 1: Initialization with IFPCA}

\begin{algorithmic}[1]

\Require Data matrix $X \in \reals^{n \times p}$, number of clusters $K$
\Ensure Initial cluster labels $\ell^{(0)}$, initial influential feature set $I^{(0)}$, p-values of KS test $ P_{KS,p}$

\State Detail for this part, please see in Alg \ref{alg:ifpca}
        


\end{algorithmic}



\vspace{1mm}
\Statex \textbf{Step 2: Iterative Loop}
\begin{algorithmic}[1]
\Require Data matrix $X \in \reals^{n \times p}$, number of clusters $K$, initial cluster labels $\ell^{(0)}$, initial influential feature set $I^{(0)}$, p-values of KS test $ P_{KS,p}$
\Ensure Predicted cluster labels $\ell$, influential feature set $I$
\State Sample $F_{\text{random}}$ as F statistics under random selected features.
\For{$t = 1, 2, \dots ,\text{max\_iter}$}
\vspace{1mm}

    \Statex \hspace{\algorithmicindent} \textbf{Step 2.1: Compute F statistic}
    \For{$j = 1$ to $p$}

        \State Computer $F^{(t)}(j)$ under $\ell^{(0)}$

        \State \parbox[t]{\dimexpr\linewidth-\algorithmicindent\relax}{%
          Normalize the F statistics with quantiles: 
          $F^{(t)}_{\text{adj}}(j) = \frac{F(j)-Q_2^d}{Q_3^d-Q_1^d}*(Q_3^t-Q_1^t)+Q_2^t$, where $Q_q^d$ and $Q_q^t$ are empirical and theoretical q-th quantiles of $F^{(t)}(j)$ and the null F-distribution $F_0$, respectively.%
        }
        
        \State $P_{F,j}^{(t)} \gets 1 - F_0(F^{(t)}_{\text{adj}}(j))$
    \EndFor
    
    \vspace{0.5mm}
    
    \Statex \hspace{\algorithmicindent} \textbf{Step 2.2 Calculate weight for F-test}
    
    
    \State For $\{1\leq m\leq p \mid m \in I^{(t-1)}\}$, $\pi_{m}^{(t-1)}=\mathrm{mean}(F^{(t)}_{\text{adj}}(m)<F_{\text{random}})$
    
    \State Sort $\pi^{(t-1)}$: $\pi_{(1)}^{(t-1)} \leq \pi_{(2)}^{(t-1)}\leq \dots \leq \pi_{(s^{(t-1)})}^{(t-1)}$, where $s^{(t-1)} = |I^{(t-1)}|$
    
    \State $ p_1^{(t)} = 1 - \exp(-e^{c - b T})$, where 
    $T = \max_{1 \leq j \leq 2s^{(t-1)}/3} \sqrt{s^{(t-1)}} \frac{j/s^{(t-1)}-\pi_{(j)}^{(t-1)}}{\sqrt{\pi_{(j)}^{(t-1)}(1-\pi_{(j)}^{(t-1)})}}$
    
    \State Weight $w^{(t)} = 1- {p_1^{(t)}}/{ (p_1^{(t)} + 0.6)}$
    
    \vspace{0.5mm}
    \Statex \hspace{\algorithmicindent} \textbf{Step 2.3: Compute core statistic}
    
    \State \parbox[t]{\dimexpr\linewidth-\algorithmicindent\relax}{ For each feature $j$, core statistic is $S_j^{(t)} = w^{(t)} \Phi^{-1}(1 - P_{F, j}^{(t)}) + (1 - w^{(t)})\Phi^{-1}(1 - P_{KS, j})$, where $\Phi^{-1}$ is inverse standard normal distribution}
    
    \vspace{1mm}
    
    \Statex \hspace{\algorithmicindent} \textbf{Step 2.4: Calculate threshold}
    \State \parbox[t]{\dimexpr\linewidth-\algorithmicindent\relax}{ For each feature $j$, $\pi_j^{(t)} = \Phi(1 - S_j^{(t)}/\sqrt{(w^{(t)})^2 + (1 - w^{(t)})^2})$, where $\Phi$ is standard normal distribution}
    
    \State Sort the $p$-values as $\pi_{(1)}^{(t)} \leq \pi_{(2)}^{(t)} \leq \cdots \leq \pi_{(p)}^{(t)}$
    
    \State The HCT can be found as $\tau^{(t)} = S_{\hat{j}}^{(t)}$, where 
    $ \hat{j} = \arg\max_{\log p \leq j \leq p/2} \frac{j/p - \pi_{(j)}^{(t)}}{\sqrt{\pi_{(j)}^{(t)}(1 - \pi_{(j)}^{(t)})}}$
    
    \State $I^{(t)} = \{1 \leq j \leq p \mid S_j^{(t)} \geq \tau^{(t)}\}$
    \vspace{1mm}
    \Statex \hspace{\algorithmicindent} \textbf{Step 2.5: Reduce dimensions and cluster}
    \State \parbox[t]{\dimexpr\linewidth-\algorithmicindent\relax}{ Apply Laplacian Eignmap or PCA on $W[:,I^{(t)}]$, retain the top $K+2$ eigenvectors to form a spectral matrix $U^{(t)}$}
    
    \State Perform $k$-means on $U^{(t)}$, then $\ell^{(t)}=k\text{-means}(U^{(t)},K)$

    \If{$r^{(t)} = \frac{\left| I^{(t)}/I^{(t-1)} \right|}{\left| I^{(t-1)} \right|}<10 \% $}
        \State \textbf{break}
    \EndIf

\EndFor
\end{algorithmic}
\end{algorithmic}
\end{algorithm}

\begin{algorithm}[H]
\caption{DeepCluster with Autoencoder and Hyperparameter Optimization}
\label{alg:deepcluster}
\begin{algorithmic}[1]
\Require Input data $X$, number of clusters $K$
\Ensure Predicted cluster labels $\ell$
\State Initialize an Optuna study to maximize clustering performance
    \For{each trial in Optuna}
    \State Sample hyperparameters: hidden size $h$, epochs $E$, iterations $T$
    \State Initialize an autoencoder model with encoder, decoder, and a classification head $\ell^{(0)}$
    \For{each iteration $t = 1$ to $T$}
        \State Encode input $W$ to get low-dimensional features $z$
        \State Normalize $z$ and apply $k$-means with $K$ clusters to obtain pseudo-labels
        \For{each epoch $e = 1$ to $E$}
            \State Decode $z$ to reconstruct input, and classify using pseudo-labels
            \State Compute total loss: reconstruction loss $+$ classification loss
            \State Update model parameters via backpropagation
            \EndFor
        \EndFor
        \State Compute silhouette score $s$ based on final cluster assignments
        \State Define objective score as: $s - 0.5 \cdot \text{final loss}$
    \EndFor
    \State Retrieve the best hyperparameters from Optuna
    \State Train the model again using the best settings, obtion predicted cluster labels $\ell$
\end{algorithmic}
\end{algorithm}

\begin{algorithm}[H]
\caption{Clustering with DEC}
\label{alg:dec}
\begin{algorithmic}[1]

\Require Input data $X$, number of clusters $K$
\Ensure Predicted cluster labels $\ell$
\State Define an autoencoder: encoder $f_\phi(x) = z$, decoder $g_\theta(z) = \hat{x}$
\For{epoch $= 1$ to $N_{\text{pretrain}}$}
    \State Compute reconstruction: $\hat{x} = g_\theta(f_\phi(x))$
    \State Minimize MSE loss: $\mathcal{L}_{\text{recon}} = \|x - \hat{x}\|^2$
\EndFor

\State Encode all data: $z = f_\phi(x)$
\State Apply $k$-means on $z$ to obtain cluster centers $\{\mu_k\}_{k=1}^{K}$
\State Initialize cluster layer with these centers
\For{epoch $= 1$ to $N_{\text{DEC}}$}
    \State Encode $z = f_\phi(x)$ and compute soft assignments $q_{ik}$:
    $q_{ik} = \frac{(1 + \|z_i - \mu_k\|^2)^{-1}}{\sum_j (1 + \|z_i - \mu_j\|^2)^{-1}}$
    \State Compute target distribution $p$:
    $p_{ik} = \frac{q_{ik}^2 / \sum_i q_{ik}}{\sum_j (q_{ij}^2 / \sum_i q_{ij})}$
    \State Minimize KL divergence loss:
    $\mathcal{L}_{\text{KL}} = \text{KL}(P \| Q) = \sum_i \sum_k p_{ik} \log \frac{p_{ik}}{q_{ik}}$
    \State Total loss: $\mathcal{L} = \mathcal{L}_{\text{KL}} + \mathcal{L}_{\text{recon}}$
    \State Update model parameters via gradient descent
\EndFor

\State Assign cluster label $\ell_i = \arg\max_k q_{ik}$ for each sample

\end{algorithmic}
\end{algorithm}

\begin{algorithm}[H]
\caption{Clustering with UMAP}
\label{alg:umap}
\begin{algorithmic}[1]
    \Require Input data $X$, number of clusters $K$
    \Ensure Predicted labels $\ell$
    \State Obtain top $K+2$ eigenvectors for $XX^{T}$, and use them as initialization for UMAP optimizer
    \State Apply UMAP on $X$, retain the top $K+2$ eigenvectors to
form a spectral matrix $U$
    \State Perform $k$-means on $U$, then predicted labels $\ell=k\text{-means}(U,K)$
\end{algorithmic}
    
\end{algorithm}

\begin{algorithm}[H]
\caption{Clustering with VAE}
\label{alg:vae}
\begin{algorithmic}[1]
\Require Input data $X$, number of clusters $K$
\Ensure Predicted labels $\ell$

\State Define encoder network $q_\phi(z|x)$ that maps input $X$ to latent mean $\mu$ and log-variance $\log \sigma^2$
\State Use reparameterization trick: $z = \mu + \sigma \cdot \epsilon$, where $\epsilon \sim \mathcal{N}(0, I)$
\State Define decoder network $p_\theta(w|x)$ to reconstruct input from latent vector

\State Train the VAE by minimizing the loss:
$
\mathcal{L}(x) = \mathbb{E}_{q_\phi(z|x)}[\log p_\theta(x|z)] - \beta \cdot D_{\text{KL}}(q_\phi(z|x) \| p(z))
$

\State Use a warm-up schedule to gradually increase $\beta$ from 0 to 1 during training

\State Encode all samples to latent space: $Z = \mu(x)$ for each $x$
\State Apply $k$-means clustering on $Z$ with $K$ clusters, obtain predicted labels $\ell$

\end{algorithmic}
\end{algorithm}

\section{IMPLEMENTATION DETAILS AND PARAMETERS}
\label{sec:implemention}
In the following subsections, we present the implementation details and hyperparameter settings of all methods involved in our study. Specifically, i-IF-Lap is our proposed iterative feature selection algorithm that incorporates Laplacian embedding to guide low-dimensional clustering. It constructs a cosine-based affinity graph and applies spectral embedding for representation learning. DeepCluster is a self-supervised deep clustering method that alternates between clustering with K-means and updating a feature encoder, which we adapt to our tabular datasets using lightweight autoencoders. IFVAE and i-IF-Lap+VAE are based on the Variational Autoencoder framework, where clustering is performed in the learned latent space, and i-IF-Lap+VAE further integrates our iterative feature selection procedure. DEC (Deep Embedded Clustering) jointly optimizes a clustering loss and a deep autoencoder, and has been used as a strong baseline for representation-based clustering. UMAP is a non-linear dimensionality reduction technique that preserves both local and global structure of the data, and we use it both as a baseline and as part of i-IF-Lap+UMAP. Each method is implemented with reasonable default parameters or carefully tuned hyperparameters, as detailed in the sections below. 

\subsection{i-IF-Lap}
In i-IF-Lap, implementation details for Laplacian Eignmap are:

Cosine distance is computed between all pairs of feature vectors in \( W[:,I^{(t)}] \).
 
The affinity matrix \( A \in \reals^{n \times s^{(t)}} \) is constructed using a Gaussian kernel applied to the cosine distances:
\[
A_{ij} = \exp(-\gamma \cdot d_{ij}^2),
\]
where \( d_{ij} \) is the cosine distance between feature vectors \( i \) and \( j \), and \( \gamma \) is a scaling parameter.

For parameters:
\begin{itemize}
  \item \textbf{Gamma (\( \gamma \)):} 1
  \item \textbf{Affinity type:} Precomputed affinity matrix
  \item \textbf{Number of nearest neighbors:} 8 
  \item \textbf{Output dimensionality:} \( K + 2 \)
  \item \textbf{Implementation:} \texttt{SpectralEmbedding} from \texttt{scikit-learn}
\end{itemize}

The resulting low-dimensional representation \( U \in \mathcal{R}^{n \times (K+2)} \) is subsequently used for clustering.

\subsection{DeepCluster}
Our dataset is relatively small and low-dimensional compared to image dataset, and therefore does not require a deep or complex neural network architecture.

For both Deepcluster and i-IF-Lap+DeepCluster, we use \texttt{optuna} to obtain optimal parameters. The parameters tuned include:
\begin{itemize}
    \item \textbf{$h$}: the size of the Autoencoder's hidden layer, selected from \{64, 128, 256\}.
    \item \textbf{$E$}: the number of training epochs per clustering iteration, ranging from 5 to 15.
    \item \textbf{$T$}: the total number of clustering-training iterations, ranging from 3 to 10.
    \item \textbf{learning\_rate}: $1 \times 10^{-3}$
\end{itemize}

\subsection{VAE}
We use the IFVAE implementation provided under the GNU GPL by Chen et al. (2023) \cite{chen2023subject}, as instructed in the repository license. The citation to the original paper is included in our manuscript. For both IFVAE and i-IF-Lap+VAE, parameters are:

\begin{itemize}
    \item \textbf{latent\_dim}: Dimensionality of the latent space in the VAE, set to 25.
    \item \textbf{batch\_size}: Mini-batch size during training, set to 50.
    \item \textbf{epochs}: Total number of training epochs, set to 100.
    \item \textbf{learning\_rate}: Learning rate used in the optimizer, set to 0.0005).
    \item \textbf{kappa}: Warm-up increment per epoch for $\beta$ in the KL divergence term, set to 1.
\end{itemize}

\subsection{DEC}
We use the publicly available DEC implementation released under the MIT License by Junyuan Xie (2015) \cite{DEC}. The license permits use, modification, and redistribution with appropriate credit. For parameters in DEC:

\begin{itemize}
    \item \textbf{Hidden layer dimensions:} A list specifying the number of neurons in the encoder and decoder layers, set to \texttt{[500, 10]} for an encoder of size $p \rightarrow 500 \rightarrow 10$ and a mirrored decoder.
    \item \textbf{Pretraining epochs ($N_{\text{pretrain}}$):} Number of epochs for unsupervised autoencoder pretraining, set to 10.
    \item \textbf{DEC training epochs ($N_{\text{DEC}}$):} Number of epochs for joint clustering optimization, set to 100.
    \item \textbf{Batch size:} Number of samples per training batch, set to 256.
    \item \textbf{Learning rate:} Learning rate for the optimizer, set to $1 \times 10^{-3}$.
\end{itemize}

\subsection{UMAP}
For both UMAP and i-IF-Lap+UMAP, parameters are:
\begin{itemize}
  \item \textbf{Number of neighbors:} 8
  \item \textbf{Metric:} Cosine distance
  \item \textbf{Embedding dimensionality:} \( K+2 \)
  \item \textbf{Angular random projection forest:} Enabled (\texttt{angular\_rp\_forest=True})
  \item \textbf{Implementation:} \texttt{umap.UMAP} from the \texttt{umap} Python package
\end{itemize}

\section{DETAILS ABOUT NUMERICAL EXPERIMENTS}

\subsection{Computer Resources}
\label{sec:computer-resources}
All experiments were conducted on Amazon Web Services (AWS) using m5.large instances. The key specifications of the compute environment are as follows:

\begin{itemize}
  \item \textbf{Instance type:} AWS m5.large
  \item \textbf{CPU:} Intel(R) Xeon(R) Platinum 8175M CPU @ 2.50GHz
  \item \textbf{Cores/Threads:} 2 cores, 4 threads (Hyperthreading enabled)
  \item \textbf{Memory:} 8 GB RAM
  \item \textbf{GPU:} None (CPU-only setup)
  \item \textbf{Virtualization:} KVM hypervisor
\end{itemize}

\subsection{Datasets}
\label{sec:app-dataset}
To facilitate comparative analysis, following tables summarize the key characteristics of the benchmark data sets used in this study. For each data set, we report three key quantities: the number of samples $n$, the number of features $p$, and the number of cluster $K$. 

We use a set of publicly available gene microarray datasets in our study. Download datasets by following links: \href{scRNA-seq}{https://data.mendeley.com/datasets/nv2x6kf5rd/1}, \href{microarray}{https://data.mendeley.com/datasets/cdsz2ddv3t/1}. These datasets are licensed under the Creative Commons Attribution 4.0 International License (CC BY 4.0), as indicated on the data hosting platform. Table \ref{microarray} includes the 8 scRNA-seq data sets, while Table \ref{RNA} contains the 10 microarray data sets.
This alignment enables consistent assessment of algorithmic performance across diverse biological contexts. 

\begin{table}[h]
\centering
\begin{tabular}{cllrrr}
\toprule
& Data name & Source & $K$ & $n$ & $p$ \\
\midrule
1 & Brain            & Pomeroy (02)               & 5 & 42  & 5,597  \\
2 & Breast cancer    & Wang et al. (05)           & 2 & 276 & 22,215 \\
3 & Colon cancer     & Alon et al. (99)           & 2 & 62  & 2,000  \\
4 & Leukemia         & Golub et al. (99)          & 2 & 72  & 3,571  \\
5 & Lung cancer (1)  & Gordon et al. (02)         & 2 & 181 & 12,533 \\
6 & Lung cancer (2)  & Bhattacharjee et al. (01)  & 2 & 203 & 12,600 \\
7 & Lymphoma         & Alizadeh et al. (00)       & 3 & 62  & 4,026  \\
8 & Prostate cancer  & Singh et al. (02)          & 2 & 102 & 6,033  \\
9 & SRBCT            & Kahn (01)                  & 4 & 63  & 2,308  \\
10 & Su cancer       & Su et al. (01)             & 2 & 174 & 7,909  \\
\bottomrule
\end{tabular}
\caption{Summary of microarray datasets with $K$ (number of clusters), $n$ (samples), and $p$ (features).}
\label{RNA}
\end{table}

\begin{table}[h]
\centering
\begin{tabular}{clrrr}
\toprule
 & Data set & $K$ & $n$ & $p$ \\
\midrule
1 & Camp1    & 7  & 777  & 13,111 \\
2 & Camp2    & 6  & 734  & 11,233 \\
3 & Darmanis & 9  & 466  & 13,400 \\
4 & Deng     & 6  & 268  & 16,347 \\
5 & Goolam   & 5  & 124  & 21,199 \\
6 & Grun     & 2  & 1,502 & 5,547  \\
7 & Li       & 9  & 561  & 25,369 \\
8 & Patel    & 5  & 430  & 5,948  \\
\bottomrule
\end{tabular}
\caption{Summary of scRNA-seq datasets with $K$ (number of clusters), $n$ (samples), and $p$ (features).}
\label{microarray}
\end{table}

For scRNA-seq data sets except Patel, we add log transformation ($X=log(X+1)$) on data matrices.

\subsection{Additional Simulation Results in Synthetic Datasets}
\label{sec:appendix-synthetic}
\begin{figure}[H]
    \centering
    \includegraphics[width=1.0\linewidth]{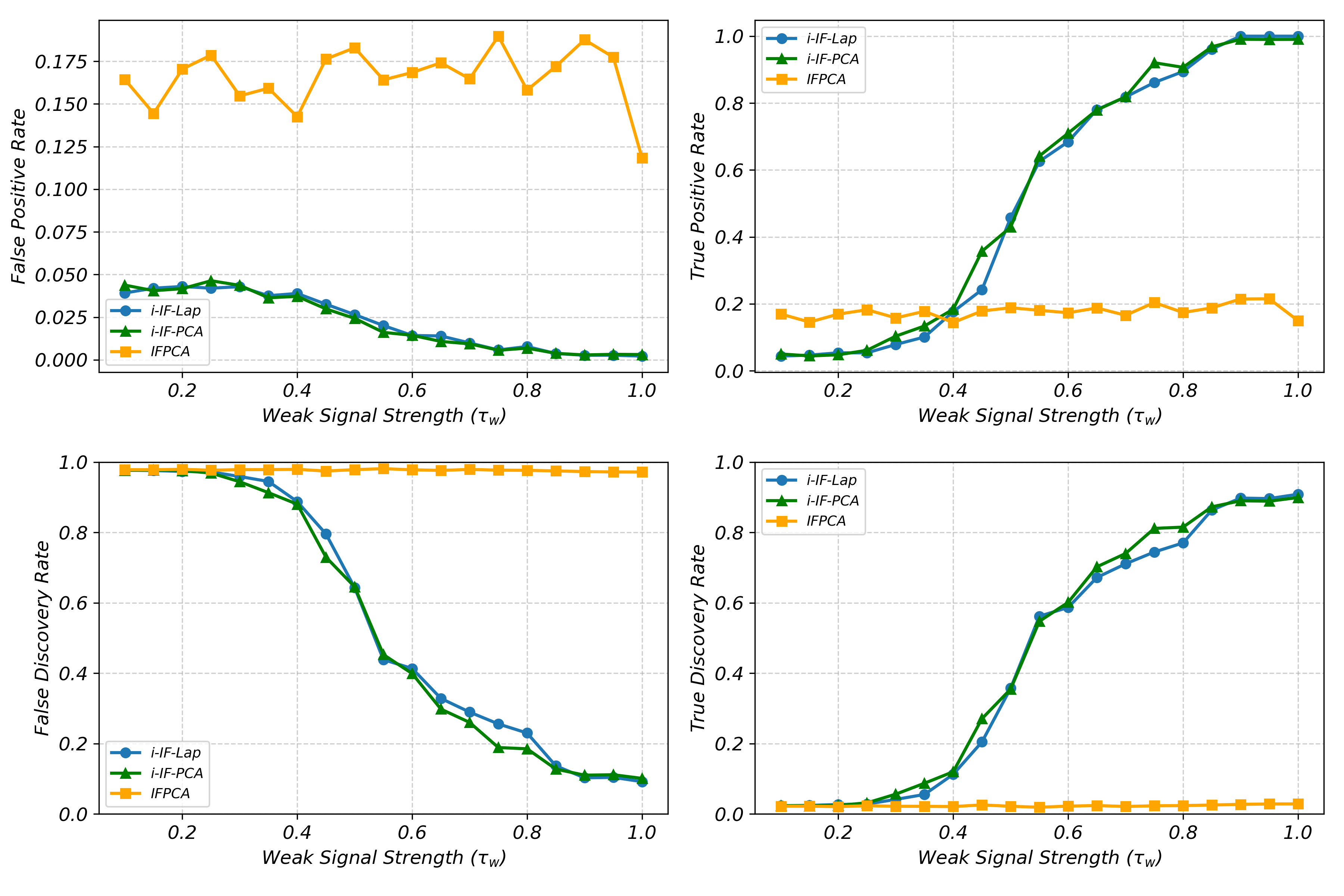}
    \caption{Comparison of feature selection performance under increasing weak signal strength ($\tau_w$). Each subplot reports a different metric: (a) False Positive Rate (FPR), (b) True Positive Rate (TPR), (c) False Discovery Rate (FDR), and (d) True Discovery Rate (TDR).}
    \label{fig:weak-signal-curves}
\end{figure}
We compare three feature selection methods: baseline IFPCA, our proposed i-IF-PCA and i-IF-Lap. As $\tau_w$ increases, we provide detailed insights into each subplot of Figure~\ref{fig:weak-signal-curves}:

\begin{itemize}
    \item \textbf{(a) False Positive Rate (FPR).} As $\tau_w$ increases, both i-IF-PCA and i-IF-Lap significantly reduce FPR, while IF-PCA maintains a consistently high FPR across all levels. This indicates that our proposed iterative methods are much more effective in suppressing noise features, especially as signal strength grows.
    
    \item \textbf{(b) True Positive Rate (TPR).} The TPR for IFPCA remains stagnant, failing to recover true signals under increasing $\tau_w$. In contrast, both i-IF-PCA and i-IF-Lap demonstrate a clear transition from low to high TPR as $\tau_w$ increases, indicating their capacity to adaptively extract true features. Notably, i-IF-Lap and i-IF-PCA achieve near-perfect TPR when $\tau_w > 0.7$.
    
    \item \textbf{(c) False Discovery Rate (FDR).} IFPCA shows extremely high FDR (close to 1), suggesting that nearly all selected features are false discoveries. Both i-IF variants show decreasing FDR as $\tau_w$ increases, with i-IF-PCA slightly outperforming i-IF-Lap under high signal strengths.
    
    \item \textbf{(d) True Discovery Rate (TDR).} TDR follows a similar trend to TPR. The iterative methods rapidly increase TDR with growing $\tau_w$, again highlighting their adaptability. i-IF-Lap and i-IF-PCA reach near-perfect TDR beyond $\tau_w = 0.8$, indicating very high fidelity in recovering true features.
\end{itemize}

\paragraph{Conclusion.}
These results further confirm that iterative frameworks (i-IF-PCA and i-IF-Lap) significantly outperform the static IFPCA in both reducing false selections and recovering weak signals, particularly when weak signals become stronger.

\section{ADDITIONAL EXPERIMENT FOR EMBEDDING}\label{sec:app-addtional}
We applied our \textit{i-IF-Learn} framework with different embedding methods on scRNA-seq datasets. We compared four popular dimensionality reduction techniques: UMAP, Autoencoder, Laplacian Eigenmap, and PCA. The clustering accuracy is reported in the form of mean (standard deviation) over 30 repetitions. 

\begin{table}[h!]
\centering
\begin{tabular}{lcccc}
\toprule
Data & UMAP & Autoencoder & Laplacian & PCA \\
\midrule
camp1    & \textbf{0.804 (0.0012)} & 0.671 (0.0418) & 0.740 (0.0000) & 0.738 (0.0000) \\
camp2    & 0.546 (0.0041) & \textbf{0.630 (0.0137)} & 0.605 (0.0000) & 0.617 (0.0000) \\
darmanis & 0.720 (0.0227) & 0.781 (0.0222) & \textbf{0.785 (0.0000)} & 0.783 (0.0000) \\
deng     & 0.626 (0.0083) & 0.861 (0.0026) & \textbf{0.869 (0.0000)} & 0.802 (0.0000) \\
goolam   & 0.665 (0.0109) & \textbf{0.916 (0.0750)} & 0.758 (0.0000) & 0.629 (0.0000) \\
grun     & 0.672 (0.0027) & 0.692 (0.0087) & \textbf{0.994 (0.0000)} & 0.991 (0.0000) \\
li       & 0.943 (0.0135) & 0.897 (0.0019) & 0.966 (0.0000) & \textbf{0.980 (0.0000)} \\
patel    & 0.946 (0.0028) & 0.771 (0.0284) & \textbf{0.942 (0.0000)} & 0.788 (0.0000) \\
\bottomrule
\end{tabular}
\caption{Clustering accuracy of different embedding methods across scRNA-seq datasets using i-IF-Learn.}
\end{table}

Overall, the results show that Laplacian Eigenmap achieves the best performance across multiple datasets, demonstrating both high accuracy and stability. While UMAP and Autoencoder sometimes achieve the highest accuracy for specific datasets, their performance is less stable, with larger standard deviations. PCA performs well on datasets with linear structure, but is generally outperformed by Laplacian Eigenmap in most other cases.

\newpage
\section{ADDITIONAL BASELINE COMPARISONS: IDC AND CLEAR}
\label{sec:appendix_idc_clear}

To further comprehensively evaluate our proposed method, we conducted additional experiments comparing i-IF-Lap against two other baseline methods: IDC and CLEAR.

First, we applied IDC to our datasets. However, we encountered out-of-memory constraints on the larger datasets, limiting its successful application to 6 datasets. As shown in Table \ref{tab:supp_idc}, our i-IF-Lap method outperforms IDC on 5 out of the 6 evaluated datasets. 

Additionally, we compared i-IF-Lap with CLEAR across 8 scRNA-seq datasets. As detailed in Table \ref{tab:supp_clear}, i-IF-Lap demonstrates superior clustering accuracy, consistently outperforming CLEAR across all 8 datasets.

\begin{table}[ht]
\centering
\caption{Clustering accuracy comparison between IDC and i-IF-Lap on 6 datasets. IDC encountered out-of-memory errors on the remaining larger datasets.}
\label{tab:supp_idc}
\footnotesize
\setlength{\tabcolsep}{4pt}
\begin{tabular}{lcccccc}
\toprule
\textbf{Method} & Brain & Colon & Lymphoma & Leukemia & Prostate & SRBCT \\
\midrule
IDC      & 0.238 & \textbf{0.645} & 0.645 & 0.653 & 0.510 & 0.524 \\
i-IF-Lap & \textbf{0.783} & 0.635 & \textbf{0.936} & \textbf{0.972} & \textbf{0.569} & \textbf{0.984} \\
\bottomrule
\end{tabular}
\end{table}

\begin{table}[ht]
\centering
\caption{Clustering accuracy comparison between CLEAR and i-IF-Lap on 8 scRNA-seq datasets.}
\label{tab:supp_clear}
\footnotesize
\setlength{\tabcolsep}{4pt}
\begin{tabular}{lcccccccc}
\toprule
\textbf{Method} & Camp1 & Camp2 & Darmanis & Deng & Goolam & Grun & Li & Patel \\
\midrule
CLEAR    & 0.597 & 0.426 & 0.487 & 0.549 & 0.742 & 0.540 & 0.886 & 0.540 \\
i-IF-Lap & \textbf{0.740} & \textbf{0.605} & \textbf{0.785} & \textbf{0.869} & \textbf{0.758} & \textbf{0.994} & \textbf{0.966} & \textbf{0.942} \\
\bottomrule
\end{tabular}
\end{table}

\section{ADDITIONAL EXPERIMENT: LASSO WITH PSEUDO-LABELS}
\label{sec:appendix_lasso}

The reviewer suggested comparing our approach with supervised feature selection methods such as Lasso. While these methods are powerful when reliable labels are available, our problem is inherently unsupervised and relies on pseudo-labels generated during the iterative procedure. As discussed in the main text, pseudo-labels in early iterations can be noisy, and supervised feature selection methods may treat these labels as ground truth, potentially leading to error propagation.

To further examine this issue, we conducted an additional experiment using Lasso-based feature selection on the microarray datasets. Specifically, we applied Lasso to the standardized data using the pseudo-labels produced by the iterative procedure, and evaluated the classification accuracy using the selected features.

\begin{table}[h]
\centering
\setlength{\tabcolsep}{2.5pt} 
\begin{tabular}{lcccccccccc}
\hline
Methods & Brain & Breast & Colon & Leukemia & Lung1 & Lung2 & Lymphoma & Prostate & SRBCT & SuCancer \\
\hline
Lasso & 0.643 & 0.627 & 0.597 & 0.931 & 0.967 & 0.783 & 0.468 & \textbf{0.618} & 0.460 & 0.500 \\
i-if-lap & \textbf{0.738} & \textbf{0.630} & \textbf{0.597} & \textbf{0.972} & \textbf{0.995} & \textbf{0.803} & \textbf{0.936} & 0.569 & \textbf{0.984} & \textbf{0.603} \\
\hline
\end{tabular}
\caption{Comparison between Lasso-based feature selection and the i-IF-Lap on microarray datasets.}
\label{tab:lasso_microarray}
\end{table}
As shown in Table~\ref{tab:lasso_microarray}, our proposed i-IF-Lap method outperforms the Lasso-based approach on 9 out of the 10 evaluated microarray datasets. Notably, in datasets such as \textit{Lymphoma} and \textit{SRBCT}, the accuracy of Lasso drops drastically compared to our framework. This substantial performance gap empirically validates our hypothesis: explicitly treating early-stage, noisy pseudo-labels as absolute ground truth---as standard supervised methods like Lasso inherently do---leads to severe error propagation. In contrast, our adaptive screening metric successfully mitigates this risk by dynamically balancing pseudo-label supervision with unsupervised signals, demonstrating the necessity and superiority of our tailored unsupervised framework.
\end{document}